\documentclass[sn-mathphys,Numbered]{sn-jnl}

\usepackage{hyperref}
\usepackage{graphicx}%
\usepackage{multirow}%
\usepackage{amsmath,amssymb,amsfonts}%
\usepackage{amsthm}%
\usepackage{mathrsfs}%
\usepackage[title]{appendix}%
\usepackage{xcolor}%
\usepackage{textcomp}%
\usepackage{manyfoot}%
\usepackage{algorithm}%
\usepackage{algorithmicx}%
\usepackage{algpseudocode}%
\usepackage{listings}%
\usepackage{enumerate}

\usepackage{microtype}
\usepackage{graphicx}
\usepackage{subfigure}
\usepackage{mathtools}
\usepackage[capitalize,noabbrev]{cleveref}
\usepackage[utf8]{inputenc} 
\usepackage[T1]{fontenc}    
\usepackage{booktabs}       
\usepackage{nicefrac}       
\usepackage{xcolor}         
\usepackage{latexsym}
\usepackage{multirow}
\usepackage{makecell}
\usepackage{enumitem}
\usepackage{graphicx}
\usepackage{setspace}
\usepackage{subfigure}
\usepackage{latexsym}
\usepackage{inconsolata}
\usepackage{bm}
\usepackage{wrapfig}
\usepackage{arydshln}
\usepackage{hyperref}
\usepackage{url}
\usepackage{pifont}
\usepackage{bbding}

\theoremstyle{thmstyleone}%
\newcommand{\cmark}{\ding{51}}%

\theoremstyle{thmstyletwo}%

\theoremstyle{thmstylethree}%

\begin{document}

\title[Towards biologically plausible computing]{Towards Biologically Plausible Computing: \\ A Comprehensive Comparison}


\author[1]{\fnm{Changze} \sur{Lv}}\email{czlv22@m.fudan.edu.cn}
\equalcont{These authors contributed equally to this work.}

\author[2]{\fnm{Yufei} \sur{Gu}}\email{yufei.gu.20@ucl.ac.uk}
\equalcont{These authors contributed equally to this work.}

\author[1]{\fnm{Zhengkang} \sur{Guo}}\email{zkguo20@fudan.edu.cn}
\equalcont{These authors contributed equally to this work.}

\author[1]{\fnm{Zhibo} \sur{Xu}}\email{zbxu23@m.fudan.edu.cn}
\equalcont{These authors contributed equally to this work.}

\author[1]{\fnm{Yixin} \sur{Wu}}\email{yixinwu23@m.fudan.edu.cn}
\equalcont{These authors contributed equally to this work.}

\author[1]{\fnm{Feiran} \sur{Zhang}}\email{frzhang23@m.fudan.edu.cn}
\equalcont{These authors contributed equally to this work.}

\author[1]{\fnm{Tianyuan} \sur{Shi}}\email{tyshi23@m.fudan.edu.cn}
\equalcont{These authors contributed equally to this work.}

\author[1]{\fnm{Zhenghua} \sur{Wang}}\email{zhenghuawang23@m.fudan.edu.cn}
\equalcont{These authors contributed equally to this work.}

\author[1]{\fnm{Ruicheng} \sur{Yin}}\email{rcyin23@m.fudan.edu.cn}
\equalcont{These authors contributed equally to this work.}

\author[1]{\fnm{Yu} \sur{Shang}}\email{yshang21@m.fudan.edu.cn}
\equalcont{These authors contributed equally to this work.}

\author[1]{\fnm{Siqi} \sur{Zhong}}

\author[1]{\fnm{Xiaohua} \sur{Wang}}

\author[1]{\fnm{Muling} \sur{Wu}}

\author[1]{\fnm{Wenhao} \sur{Liu}}

\author[1]{\fnm{Tianlong} \sur{Li}}

\author[1]{\fnm{Jianhao} \sur{Zhu}}

\author[1]{\fnm{Cenyuan} \sur{Zhang}}

\author[1]{\fnm{Zixuan} \sur{Ling}}

\author*[1]{\fnm{Xiaoqing} \sur{Zheng}}\email{zhengxq@fudan.edu.cn}


\affil[1]{\orgdiv{Department of Computer Science}, \orgname{Fudan University}}

\affil[2]{\orgdiv{Department of Computer Science}, \orgname{University College London}}


\abstract{
Backpropagation is a cornerstone algorithm in training neural networks for supervised learning, which uses a gradient descent method to update network weights by minimizing the discrepancy between actual and desired outputs. 
Despite its pivotal role in propelling deep learning advancements, the biological plausibility of backpropagation is questioned due to its requirements for weight symmetry, global error computation, and dual-phase training.
To address this long-standing challenge, many studies have endeavored to devise biologically plausible training algorithms. 
However, a fully biologically plausible algorithm for training multilayer neural networks remains elusive, and interpretations of biological plausibility vary among researchers.
In this study, we establish criteria for biological plausibility that a desirable learning algorithm should meet.
Using these criteria, we evaluate a range of existing algorithms considered to be biologically plausible, including Hebbian learning, spike-timing-dependent plasticity, feedback alignment, target propagation, predictive coding, forward-forward algorithm, perturbation learning, local losses, and energy-based learning.
Additionally, we empirically evaluate these algorithms across diverse network architectures and datasets. We compare the feature representations learned by these algorithms with brain activity recorded by non-invasive devices under identical stimuli, aiming to identify which algorithm can most accurately replicate brain activity patterns.
We are hopeful that this study could inspire the development of new biologically plausible algorithms for training multilayer networks, thereby fostering progress in both the fields of neuroscience and machine learning.
}


\maketitle

\section{Introduction}\label{sec:intro}


Backpropagation \cite{rumelhart1986learning} has been instrumental in the rapid development of deep learning \cite{lecun2015deep}, establishing itself as the standard approach for training neural networks in supervised learning settings.
This algorithm leverages a gradient descent method to iteratively adjust network weights, thereby minimizing the errors between the actual outputs of the network and the desired outputs.
Despite its undeniable success and widespread adoption in various applications ranging from image recognition \cite{Delorme2001NetworksOI} to natural language processing \cite{Devlin2019BERTPO,brown2020language}, the biological plausibility of backpropagation remains a subject of intense debate among researchers in both neuroscience and computational science \cite{bianchini1997terminal,payeur2021burst,zahid2023predictive}.

The primary criticisms of backpropagation's biological plausibility stem from several unrealistic requirements: the symmetry of weight updates in the forward and backward passes \cite{stork1989backpropagation}, the computation of global errors that must be propagated backward through all layers \cite{crick1989recent}, and the necessity of a dual-phase training process involving distinct forward and backward passes \cite{lillicrap2020backpropagation}.
These features are not only computationally intensive but also lack clear analogs in neurobiological processes, which operate under constraints of local information processing and low energy consumption.

Recognizing these limitations, the research community has made significant strides toward developing alternative training algorithms that could potentially align more closely with biological processes.
Efforts have ranged from revisiting classical theories such as Hebbian learning \cite{munakata2004hebbian} to exploring newer concepts like spike-timing-dependent plasticity (STDP) \cite{Song2000CompetitiveHL} and feedback alignment \cite{Lillicrap2016RandomSF}.
Each of these approaches offers a unique perspective on how synaptic changes might occur in a biologically plausible manner, yet a consensus on a fully effective and biologically accurate training method remains out of reach.

In this paper, we aim to critically assess the current landscape of what are considered biologically plausible learning algorithms.
We begin by establishing a set of criteria (See Section \ref{sec:preliminary}) that any algorithm must meet to be considered biologically plausible.
These criteria are designed to encapsulate essential aspects of neurobiological learning, such as the locality of computations, the absence of a global error signal, and energy efficiency in synaptic adjustments.

Following the establishment of these criteria, we embark on a comprehensive evaluation of various learning algorithms that have been proposed in the literature as biologically plausible models.
This includes but is not limited to, Hebbian learning \cite{munakata2004hebbian}, STDP \cite{Song2000CompetitiveHL}, feedback alignment \cite{Lillicrap2016RandomSF}, target propagation \cite{Bengio14}, predictive coding \cite{rao1999predictive}, the forward-forward algorithm \cite{hinton2022forward}, perturbation learning \cite{Williams92,WerfelXS03}, local losses \cite{MarblestoneWK16}, and energy-based learning \cite{hopfield1984neurons,scellier2017equilibrium}.
Our evaluation not only examines the theoretical foundations and computational efficiency of these algorithms but also involves empirical assessments across various neural network architectures and datasets.

Furthermore, we extend our analysis to include a comparative study of the feature representations learned by these algorithms against actual brain activity patterns.
This was achieved by using non-invasive brain recording techniques, such as fMRI and EEG, to record neural responses to identical stimuli and compare these responses to the activations within artificial neural networks trained by the aforementioned algorithms.

By providing a thorough analysis of these algorithms and their ability to model brain-like learning processes, we aspire to contribute to the ongoing dialogue between the fields of neuroscience and artificial intelligence.
Ultimately, we hope that this study will not only shed light on the current capabilities and limitations of proposed biologically plausible learning algorithms but also inspire further research and development in this crucial area.
This endeavor aims to bridge the gap between biological learning processes and artificial learning systems, paving the way for the development of more efficient, robust, and biologically inspired computational models.

\section{Criteria for Biological Plausibility}
\label{sec:preliminary}
As the field of artificial intelligence strives to develop algorithms that are not only efficient but also mimic the fundamental mechanisms of human cognition, the concept of biologically plausible computing has gained significant interest.
Traditional backpropagation, while effective for training deep neural networks, diverges from known biological processes in several key areas. This divergence has prompted researchers to explore alternative algorithms that might adhere more closely to the principles observed in natural neural systems.
However, the notion of biological plausibility is not universally defined and varies significantly across different studies.
To navigate this complexity, it is essential to establish clear criteria against which these models can be evaluated.
In this section, we outline these criteria and assess existing models accordingly.

We propose five criteria for evaluating the biological plausibility of learning algorithms, summarized from existing literature.

\begin{enumerate}[]

\item \textbf{Asymmetry of Forward and Backward Weights}: In conventional neural networks, the forward-path neurons transmit their synaptic weights to a feedback path, a process known as weight transport, which is biologically implausible.
Real neurons are unlikely to share precise synaptic weights in such a manner.

\item \textbf{Local Error Representation}: Biological synapses are believed to modify their strength based on local information, without access to a global error signal.
This contrasts with the gradient descent approach where the direction of the error gradient is typically computed using global information.

\item \textbf{Non-parallel Training (or Non-Two-Stage Learning)}: Traditional training methods often require a clear distinction between the phases of forward and backward propagation, which is not a feature of biological learning.
Bio-plausible methods are explored for their ability to simplify the learning process into more continuous, possibly overlapping phases that better mimic biological learning dynamics.

\item \textbf{Models of Neurons}: The majority of artificial neural networks utilize neurons that output continuous values, intended to represent the firing rates of biological neurons, which in reality use spikes. This discrepancy is addressed through models that incorporate more realistic, spiking neuron models and learning rules adapted to such models.

\item \textbf{Unsigned Error Signals}: In biological systems, error signals are not typically signed or extreme-valued as in many artificial systems. Some learning rules attempt to approximate the way biological systems might handle error feedback without relying on these artificial constructs.

    
\end{enumerate}

\section{Brain-Inspired Learning Algorithms}\label{sec:algo}

\begin{figure}
\centering
\includegraphics[width=0.99 \textwidth]{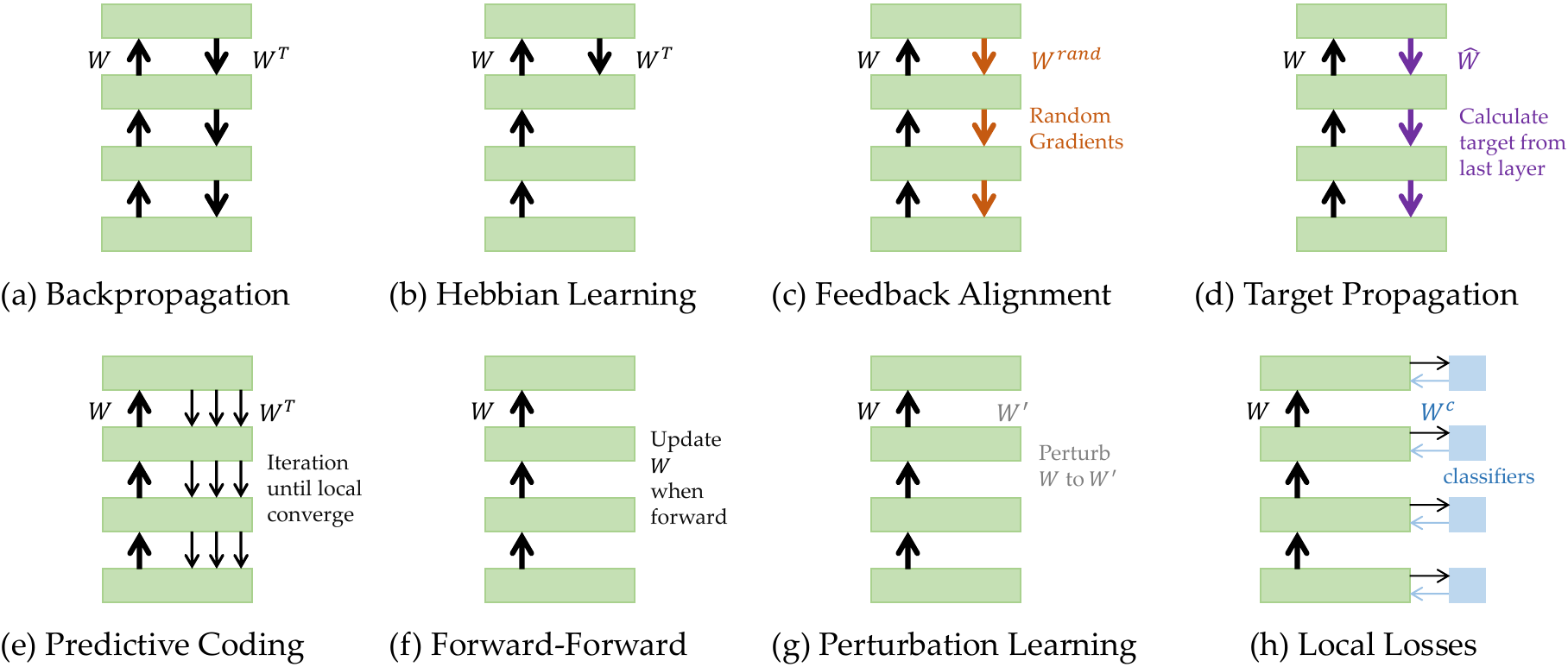}
\caption{Illustrations of various brain-inspired learning algorithms. (a) Classic backpropagation; (b) In Hebbian learning, the final classification layer is trained using gradients; (c) In feedback alignment, the weight matrix \( W \) is replaced with a random matrix during backpropagation; (d) In target propagation, two sets of weights are used: forward weights \( W \) and backward weights \( \hat{W} \), with \( \hat{W} \) used to calculate targets from the final layer; (e) In predictive coding, the transposed weights \( W^T \) are used iteratively for local convergence; (f) The forward-forward algorithm updates weights during the forward pass; (g) In perturbation learning, the weight \( W \) is randomly perturbed after the forward pass, generating a new \( W' \) for the next iteration; (h) In local losses, classic backpropagation is applied layer by layer.}
\label{fig:algo}
\end{figure}

In this section, we review nine representative brain-inspired learning algorithms in detail, including Hebbian learning, spike-timing-dependent plasticity, feedback alignment, target propagation, predictive coding, forward-forward algorithm, perturbation learning, local losses, and energy-based learning method.
Moreover, we show the illustration of these methods in Figure \ref{fig:algo}.

\subsection{Hebbian Learning}
Hebbian learning is a fundamental concept in neural network models aimed at explaining how neurons in the brain adapt and learn from experience \cite{munakata2004hebbian}. Proposed by \cite{hebb1949organisation}, the theory suggests that the efficiency of a neuron in contributing to the firing of another neuron can increase if the two neurons are repeatedly involved in each other's activation. The theory is often summarized as ``Cells that fire together wire together.'' It is biologically plausible and ecologically valid, relying solely on inputs entering the system to produce patterns of activity, without explicit tasks or teaching signals\cite{doi:10.1126/science.1372754}.

In neural network models, Hebbian learning is implemented through changes in the strength of connection weights between units. The strength of a connection weight determines the efficacy of a sending unit in activating a receiving unit. Through Hebbian learning, weights change as a function of the activity levels of the units involved. The basic form of the Hebbian learning rule can be expressed as:
\begin{equation}
   \label{eq:hebbian_rule_1}
\Delta w = \eta y(x,w) x 
\end{equation}
where $\Delta w$ represents the weight change, $y(x,w)$, a function of the input and the weights, is the post-synaptic activation of the neuron $\eta$, determines the speed at which the weights change in response to unit activation. The activation level of a unit typically falls within the range of 0 to 1 and is calculated as a nonlinear function of the activation of other units and the strength of their connections to the unit.

The main problem of rule 1.1 is that it only allows weights to grow, not decrease. To prevent the weight vector from growing unbounded, it is possible to normalize it after every update, introducing a weight decay term proportional to the value of the weight, given by:
\begin{equation}
   \label{eq:hebbian_rule_2}
\Delta w = \eta  y(x,w) (x - w) 
\end{equation}

Equation \eqref{eq:hebbian_rule_2} has the physical interpretation that at each iteration, the weight vector is modified by taking a step towards the input, with the size of the step being proportional to the similarity between the input and the weight vector. Consequently, if a similar input is presented again in the future, the neuron will be more likely to produce a stronger response. If an input (or a cluster of similar inputs) is repeatedly presented to the neuron, the weight vector tends to converge towards it, eventually acting as a matching filter. In other words, the input is memorized in the synaptic weights. From this perspective, the neuron can be seen as an entity that, when stimulated with a frequent pattern, learns to recognize it.

The abstract nature of the Hebbian learning rule allows it to be applicable to different types of neural networks and learning tasks. There can be multiple methods to implement Hebbian learning.
HebbNet\cite{gupta2021hebbnet} is a neural network that utilizes an improved Hebbian approach with updated activation thresholds and gradient sparsity. SoftHebb\cite{moraitis2022softhebb} is a variant that combines standard deep learning elements with a Hebbian-like plasticity. It maintains a Bayesian generative model of the data without supervision and minimizes cross-entropy, offering advantages beyond traditional Hebbian efficiency.

\subsection{Spike-timing-dependent Plasticity}

In biological neural networks, signals are transmitted between neurons through spikes. The pre-synaptic neuron, responsible for signaling, generates a spike, releasing neurotransmitters within the synapse—a structure connecting it to the post-synaptic neuron, which, in turn, triggers the rise of the postsynaptic potential. As the potential accumulates and surpasses a certain threshold, a spike occurs in the postsynaptic neuron. The efficacy of pre-synaptic excitement in inducing post-synaptic excitement serves as the synaptic weight between two layers of neurons.  

When a post-synaptic spike closely follows a pre-synaptic spike, it implies a strong association between the excitations of the two neurons, suggesting a causal relationship. In such cases, it is recommended to strengthen the synaptic weight between them. The shorter the interval between the spikes, the stronger the inferred causal relationship, leading to a greater increase in weight. Conversely, if the post-synaptic spike precedes the pre-synaptic spike, the intuitive assumption is that there is no causal relationship, resulting in a reduction in the synaptic weight. Spike-time-dependent plasticity (STDP), a learning rule within Hebbian learning models, adjusts synaptic weights based on the timing of spikes between pre- and post-synaptic neurons at the synapse, aligning with the aforementioned approach.

As proposed by \cite{Song2000CompetitiveHL}, pre-synaptic excitation preceding post-synaptic excitation leads to long-term potentiation (LTP), resulting in an increased synaptic weight. Conversely, when pre-synaptic excitation follows post-synaptic excitation, it induces long-term depression (LTD), causing a decrease in synaptic weight. The magnitude of weight change is contingent upon the timing difference between spikes. This phenomenon, initially observed in neuroscience experiments, has been formulated as a learning rule for neuronal synapses. The mathematical expression for this rule is:
\begin{equation}
\Delta w= \left\{
\begin{aligned}
& \alpha_+ e^{\frac{\Delta t}{\tau_+}} \text{ , if } \Delta t>0\\
& \alpha_- e^{\frac{\Delta t}{\tau_-}} \text{ , if } \Delta t<0
\end{aligned}
\right.
\label{Class_stdp}
\end{equation}
Where the $\Delta t=t_{post}-t_{pre}$ represents the time difference of spikes between the two neurons at the synapse. $\alpha_+$ and $\alpha_-$ determine the maximum amounts of synaptic modification and $\tau_+$ and $\tau_-$ are the time constants.

Spike-timing-dependent plasticity (STDP) manifests across a variety of species. Following an initial discussion of STDP, we transition to an exploration of its associated biological mechanisms, particularly the $Ca^{2+}$ hypothesis\cite{artola1993long,lisman1989mechanism}. Notably, a high influx of $Ca^{2+}$ in the post-synaptic neuron induces Long-Term Potentiation (LTP), while a moderate influx results in Long-Term Depression (LTD). Distinct $Ca^{2+}$ levels activate specific molecular pathways; high $Ca^{2+}$ activates CaMKII for LTP, while moderate levels activate PP1 and calcineurin for LTD\cite{caporale2008spike}. The level of $Ca^{2+}$ influx hinges on the interval between the excitatory postsynaptic potential (EPSP) triggered by the pre-synaptic neuron's activation and the backpropagating action potential (BAP) initiated by the post-synaptic neuron's activation. Positive intervals signify BAP occurring shortly after EPSP, while negative intervals denote BAP preceding EPSP. When the interval is positive, the BAP leads to $Mg^{2+}$ unblocking NMDA receptors\cite{Kampa_Clements_Jonas_Stuart_2004}. Simultaneously, EPSP deactivates specific ion channels, augmenting BAP magnitude\cite{Watanabe_Hoffman_Migliore_Johnston_2002, Hoffman_Magee_Colbert_Johnston_1997}. This intensified BAP activates voltage-dependent $Ca^{2+}$ channels (VDCCs)\cite{Bi_Poo_1998}. The interplay between EPSE and BAP in positive intervals results in a substantial $Ca^{2+}$ influx in the post-synaptic neuron. Conversely, when the interval is negative, the afterdepolarization of BAP coinciding with EPSP induces a moderate $Ca^{2+}$ influx\cite{caporale2008spike}. BAP, causing $Ca^{2+}$ influx through VDCCs, inhibits NMDA receptors\cite{Rosenmund_Feltz_Westbrook_1995}. These interactions between BAP and EPSP in negative intervals lead to a moderate level of $Ca^{2+}$.

The STDP learning rule has many different mathematical forms and boasts numerous variations.  Rather than computing $\Delta w$ between the current spike and every single spike in the past when a spike occurs at the synapse, a more efficient approach involves maintaining a synapse trace for both pre- and post-synaptic neurons. These traces undergo continuous modification over successive timesteps:
\begin{equation}
\begin{aligned}
 x_{pre}^{(t_i)}=&\lambda x_{pre}^{(t_{i-1})}+s_{pre}^{(t_i)} \\
 x_{post}^{(t_i)}=&\lambda x_{post}^{(t_{i-1})}+s_{post}^{(t_i)} \\
 \Delta w=&s_{post}^{(t_i)} x_{pre}^{(t_i)}-s_{pre}^{(t_i)} x_{post}^{t_{(i-1)}}
\end{aligned}    
\end{equation}
Where $t_{i-1}$ and $t_i$ are two adjacent timesteps. $x$ stands for the synapse trace and decays over time by a coefficient $\lambda$ ($0<\lambda<1$). $s^{(t_i)}=1$ when a spike occurs at $t_i$, and $s^{(t_i)}=0$ if no spike occurs.

Numerous studies on the effects of Spike-Timing-Dependent Plasticity (STDP) through virtual neural network simulations have been conducted. Research indicates that STDP confers selectivity to neurons towards different input signals \cite{Delorme2001NetworksOI} and fosters synchronous neuron activities \cite{Zhigulin2004AnIR}\cite{BofilliPetit2004SynchronyDA}. Notably, the weights between pre- and post-synaptic neurons, influenced by STDP, typically exhibit a bimodal distribution. Moreover, STDP demonstrates the ability to maximize mutual information between input and output neuron spikes \cite{Toyoizumi2004SpiketimingDP}. Applying the STDP learning rule enables a single neuron to detect a single hidden repeating spatiotemporal pattern \cite{Masquelier2008SpikeTD}, while multiple neurons can selectively detect different patterns \cite{Masquelier2009CompetitiveSS} from a large set of spike trains with inherent noise. This suggests that STDP serves as an unsupervised learning algorithm with the capability to handle time-series data.

Significant efforts have been devoted to applying the STDP algorithm, primarily through spiking neural networks, to various machine learning tasks, particularly in the field of image processing. Experimental datasets include well-known sets such as MNIST, CIFAR-10, Caltech face, and motorbike datasets. Diverse neural network architectures have been employed, ranging from hierarchical networks inspired by the human visual cortex \cite{Beyeler2013CategorizationAD}\cite{Masquelier2007UnsupervisedLO}\cite{Kheradpisheh2015BioinspiredUL}, single-layer fully connected networks with inhibitory neurons \cite{Diehl2015UnsupervisedLO}\cite{Querlioz2013ImmunityTD}, to deep convolutional networks \cite{Kheradpisheh2016STDPbasedSD} and liquid state machines \cite{Ivanov2021IncreasingLS}. It is noteworthy that certain techniques are commonly employed across these studies. For instance, preprocessing of the original image, including the use of neuromorphic datasets and the application of Difference of Gaussian filters, specific receptive field configurations, latency encoding, and lateral inhibition, have been integral components in many works.

In practical experiments, we usually use 
 equation\ref{Class_stdp} to apply STDP, but this learning rule only be found in a small wide brain area in specific environments. There are two classes of neurons which can be called excitatory and inhibitory neurons playing an important role in the processes of learning, the STDP between them are apparently different, additionally, STDP between the neurons from the same position of the brain in different species also be different\cite{Tzounopoulos_Rubio_Keen_Trussell_2007}. the trigger patterns of STDP in vivo are also different from the binary pair pattern from equation\ref{Class_stdp}\cite{Softky_Koch_1993}, such as the different frequency burst to apply in pre-synaptic neuron, the triple pairs of spikes between pre-synaptic neuron and post-synaptic neuron and the pairs of burst and spike. The STDP in vivo also be influenced by the dendritic position which the post-synaptic membrane is in\cite{Rao_Sejnowski_2001}. In addition to the spiking of the pre-and postsynaptic neurons, STDP is also regulated by other inputs. In particular, neuromodulators and inhibitory activity in the network can affect both the magnitude and the temporal window of STDP\cite{Bear_Singer_1986}. So, these phenomenons in vivo prove that the STDP we use might not fit in the biological plausibility.

\subsection{Feedback Alignment}

The concept of Feedback Alignment (FA), originally proposed by Lillicrap et al. \cite{Lillicrap2016RandomSF}, presents an intriguing paradigm in neural network training. One of FA's remarkable attributes lies in its resolution of the weight transport problem, as elucidated in \cite{Grossberg1987CompetitiveLF}, aligning it more closely to the mechanisms observed in the human brain. Subsequent research derived from FA has not only facilitated the localization of error signal transmission \cite{Nkland2016DirectFA}, but also delved into aspects like update locking \cite{Frenkel2021LearningWF}, further enhancing its biological plausibility.

Diverging from conventional backpropagation methods, FA shows that one can achieve similar effects and accuracy on classification tasks by replacing feedback weights with fixed random synaptic weights. Central to this theory is the notion that the precise alignment of feedback weights to the transpose of the forward weights $W^T$ is unnecessary during training. Instead, a designated weight matrix $B$ suffices in steering the network in roughly the same direction as backpropagation, thereby facilitating network training. The replacement matrix $B$ chosen only has to ensure the fulfillment of the following equation on an average basis:

    \begin{equation}
        e^T W Be>0
    \end{equation}

Where $e$ denotes the error of the network’s output, and $W$ represents the synaptic weights of the forward path. However, studies \cite{Liao2015HowII} have revealed a notable decline in FA’s performance when employing deeper convolutional architectures, in contrast to its comparable performance to BP in simpler MLP networks.

Direct Feedback Alignment (DFA) \cite{Nkland2016DirectFA} , an offshoot of FA, emerges as a prominent avenue of research aimed at addressing this limitation. DFA tackles this challenge by introducing direct feedback paths and distinct fixed random weights for each hidden layer. This strategic design localizes learning signals by establishing direct connections between errors and individual layers, bypassing the conventional layer-by-layer backpropagation from the output. Consequently, training deeper networks becomes more feasible, attributed to the disentangled feedback paths offering increased flexibility in transmitting error signals. Learning is thus regarded as an extension of a forward pass, marking DFA as a notable stride towards biological plausibility.

To formalize it, traditional FA updates previous hidden layers based on the change of the following layer, depicted as:

    \begin{equation}
        \delta a_2=(B_2e) \odot f^\prime(a_2), \delta a_1=(B_1 \delta a_2) \odot f^\prime(a_1)
    \end{equation}

While DFA directly updates each hidden layer in accordance to distinct weight matrices and the error of the output layer:

    \begin{equation}
        \delta a_2=(B_2e) \odot f^\prime(a_2), \delta a_1=(B_1 e) \odot f^\prime(a_1)
    \end{equation}

Here $B_i$ represents a fixed random weight matrix, $\odot$ indicates element-wise multiplication, and $f^\prime$ denotes the derivative of the non-linearity function of hidden layers.

Empirical evidence underscores the efficacy of DFA, notably in reducing training time and narrowing the accuracy gap compared to BP on datasets like MNIST and CIFAR \cite{Frenkel2021LearningWF}. Owing to its simplicity and effectiveness, DFA is widely adopted as a foundational model in subsequent research endeavors. Nevertheless, experimental outcomes \cite{Bartunov2018AssessingTS} remain somewhat limited when scaling to more intricate network architectures or larger datasets like ImageNet.

\subsection{Target Propagation}

The backpropagation algorithm lacks biological plausibility, as in the human brain, biological neurons are interspersed with linear and nonlinear elements\cite{Grossberg_1987}. Utilizing backpropagation through feedback paths for the propagation of credit assignment necessitates precise knowledge of the nonlinear derivatives employed in the corresponding feedforward computations. Additionally, it requires alternating between exact feedforward propagation and backpropagation processes across different neuronal layers. This mechanism of gradient communication and weight transfer is biologically impractical\cite{Lillicrap_Santoro_Marris_Akerman_Hinton_2020}.

Inspired by earlier research, Bengio et al. \cite{Bengio_2014} proposed Target Propagation (TP) as a novel credit assignment approach in response to the challenges posed by the backpropagation method. TP assigns a target value $\hat{h_l}$ to each layer $l$, rather than employing a loss gradient. These target values are designed to be in close proximity to the activation values, with the potential to yield a reduced loss if achieved during the feedforward phase.

A distinct aspect of Target Propagation\cite{Lee_Zhang_Fischer_Bengio_2014} is that its backward pass operates within the same dimensional framework as the forward-pass neural activities. The objective in this phase is to align the layer activities with those induced backward, thereby facilitating the generation of the desired output. Upon receiving an input, the final output layer undergoes feedforward propagation and is directly optimized to minimize the loss. In contrast, the remaining layers are oriented towards aligning with their assigned target values. The training process involves two types of losses at each layer level. Inverse loss is used to train an approximate inverse, which is parameterized in a manner analogous to the forward computation:
    \begin{equation}
        \mathfrak{L}_l^{inv}(\lambda _l)\space =\space \left \| h_{l-1}-g(f(h_{l-1};\theta _{l-1});\lambda _l) \right \|_2^2 
    \end{equation}
where $g$ is the approximate inverse: $g(h_l;\lambda _l)=\sigma _l(V_lh_l+c_l),\lambda _l=\{V_l,c_l\}$.
Forward loss imposes a penalty on layer parameters that result in activations divergent from their designated targets:
    \begin{equation}
        \mathfrak{L}_l(\theta _l)=\left \| f(h_l;\theta _l)-\hat{h}_{l+1}  \right \|_2^2  
    \end{equation}
These losses are localized, impacting only the parameters of the individual layer, and do not take into account any implicit dependencies on the parameters of other layers.

Vanilla Target Propagation calculates targets by back-propagating the targets from higher layers through layer-specific inverses:
    \begin{equation}
        \hat{h}_l=g(\hat{h}_{l+1};\lambda _{l+1}) 
    \end{equation}
However, this simplistic approach may encounter difficulties in scenarios where different instances of the same class present diverse appearances. In such cases, TP tends to enforce uniformity in their representations across all layers, including the early ones. To address this issue, the Difference Target Propagation (DTP) was introduced, incorporating linear correction terms into the feedback process:
    \begin{equation}
        \hat{h}_l=g(\hat{h}_{l+1};\lambda _{l+1})+[h_l-g(h_{l+1};\lambda _{l+1})]
    \end{equation}
The second term is the reconstruction error, providing a linear stabilizer for the inaccuracies in inverse functions. This enhancement significantly improves the recognition performance of the Target Propagation method. In the seminal study by Lee et al. \cite{Lee_Zhang_Fischer_Bengio_2014}, the target for the penultimate layer was determined using network loss gradients, deviating from standard TP methods. \cite{Bartunov_Santoro_Richards_Marris_Hinton_Lillicrap_2018} propose the Simplified DTP (SDTP) as a refinement to DTP, where the target for the penultimate layer is computed according to eq.10. This modification effectively eliminates the biologically unrealistic aspects of gradient communication and associated weight-transport in the TP algorithm.

\subsection{Predictive Coding}
Originally proposed by \cite{rao1999predictive}, predictive coding (PC) is an influential theory in computational and cognitive neuroscience. The central idea of the theory is that the brain is composed of a hierarchy of layers, while high layers predict the activities of adjacent low layers. The entire brain maintains a cognitive model of the world, activities that cannot be accurately predicted are regarded as prediction errors which will be transmitted upwards for high-level to process. Over time, the synaptic connections between high and low levels are updated until the prediction error of the entire system is minimized.
While predictive coding originated in theoretical neuroscience as a model of information processing in the cortex, recent work has developed the idea into a general-purpose algorithm able to train neural networks using only local computations.
Compared to the classical backpropagation algorithm, hierarchical predictive coding models are considered to be more biologically plausible. When Rao and Ballard first proposed the hierarchical predictive coding model in 1999, they cautiously suggested the biological interpretability of predictive coding by indirectly highlighting similarities between the hierarchical predictive coding model and classical effects such as endstopping in the visual processing regions of the brain cortex \cite{rao1999predictive}.
Since then, the emergence of predictive coding theory has, to some extent, challenged cognitive neuroscience. Consequently, a substantial amount of neuroanatomical research has been conducted to validate or refute predictive coding. This research includes a wealth of anatomical and physiological evidence supporting predictive coding as an assumption for information transmission within the cortical hierarchy, especially in early visual processing \cite{friston2008hierarchical}.
However, despite abundant indirect evidence suggesting the potential existence of predictive coding mechanisms in the cortical regions of the brain, crucial direct evidence is still lacking. Caution should be exercised in assessing the biological plausibility of predictive coding.

Hierarchical predictive coding networks (PCNs) composed of $L$ hidden layers include two kinds of neurons.
 $x_{l,t}$ denotes the neuron that encodes time-depend predictions in layer $l$ at time $t$, $\epsilon_t$ denotes the neuron that compute prediction errors in layer $l$ at time $t$($l \in\{0,...L-1\}$). In such cases, the input signal is transmitted from low-level to high-level, and high-level neurons predict the value from the following layer according to:
\begin{equation}
    \mu_{l,t}=\theta_{l+1}f(x_{l+1,t})
\end{equation}
where $f$ is a nonlinear function, and $\theta_{l+1}$ denotes the matrix of weights connecting layer $l+1$ to layer $l$. Prediction error represents the difference between actual activity and its prediction, which is denoted by $\epsilon_{l,t}=x_{l,t}-\mu_{l,t}$. The errors are then propagated down the hierarchy and used in the learning process to update the weights of the network. Ultimately, the learning algorithm optimizes a global energy function, defined as the sum of squared prediction errors at each layer:
\begin{equation}
   \mathcal{F}=\frac{1}{2}\sum_{l=0}^{L-1}||\epsilon_{l,t}||^2
\end{equation}
\par
During training, the highest layer is fixed to an input data point, and the lowest layer is fixed to a label or target vector. During a process called inference, the weight parameters are fixed, and the neural activities are continuously updated to minimize the energy function by running gradient descent until convergence, at which point a single weight update is performed. During the weight update, the value nodes are fixed, and the weight parameters are updated via gradient descent on the same energy function. When defining inference and weight update this way, every computation only needs local information to be updated \cite{whittington2017approximation}.

During testing, only the lowest layer is fixed to the data, so the network infers the label given a test point. This process is equivalent to the inference phase described above: the weight parameters are fixed, and the neural activities are updated until convergence by running gradient descent on the energy function. Note that different works follow different paradigms for the order of updating the neural activities $x_t$ and weights $\theta_t$. In most works, neural activities are all simultaneously updated for T iterations with the aim of reaching convergence of the inference process, and the weights are updated once upon convergence \cite{millidge2022predictive,whittington2017approximation,buckley2017free}. However, in certain cases, it has been noted that updating the weights and activities of different layers in different moments yields a better performance \cite{han2018deep,ororbia2022neural}.

PCNs can be mathematically derived as variational inference on hierarchical Gaussian generative models. The hierarchical model consists of multiple layers indexed by $l$.The distribution of activations at each layer can be approximated as Gaussian distribution with a mean given by a nonlinear function $f(x_{l+1})$ with parameters $\theta_{l+1}$ and an identity covariance $I$,
    $$p(x_{0,t}... x_{L,t})=p(x_{L,t})\prod_{l=0}^{L-1}p(x_{l,t}|x_{l+1,t})$$
    $$p(x_{l,t}|x_{l+1,t})=\mathcal{N}(x_{l,t};\theta_{t+1}f(x_{l+1,t}),I)$$
For example, consider a scenario where the input layer $l=0$ is held constant at a specific data item$s_{in}$. Our objective is to deduce the state of the remaining network based on this conditioning $p(x_{0},...,x{L-1}|x_{L})$. Solving this inference problem can be accomplished through variational inference. Broadly speaking, variational inference tackles the challenge of approximating an intractable inference by framing it as an optimization problem. This involves optimizing the parameters of an approximate variational posterior distribution $q$ to minimize its divergence from the optimal posterior $p$.The optimization process revolves around minimizing an upper bound on this divergence, referred to as the variational free energy $\mathcal{F}_t$.
In PCNs, our assumption is that the variational posterior is factorized into independent posteriors for each layer $q(x_{0,t},...,x_{L-1,t})=\prod \limits_{l=0}^{L-1}q(x_{l,t})$. When combined with the Laplace approximation, this simplification enables us to express the free energy as a sum of squared prediction errors.
\begin{equation}
    \mathcal{F}_t\simeq\sum_{l=0}^{L-1}\log_{p(x_{l,t}|x_{l+1,t})}\simeq\sum_{l=0}^{L-1}||\epsilon_{l,t}||^2
\end{equation}

As we mentioned above, $\epsilon_{l,t}=x-\theta_{l+1} f(x_{l+1,t})$ is the prediction error for each layer.When applied to ANNs, we generally operate under the assumption that the dependencies between layers are characterized by a parameter matrix $\theta_{l+1}$. This matrix corresponds to the weights within an ANN. Consequently, updates to both the activations $x_{l,t}$ and weights can be performed through gradient descent on the free energy.
\begin{equation}
    \mathrm{d}x_{l}/\mathrm{d}t \propto -\partial \mathcal{F}_t/\partial x_{l,t}
\end{equation}

\begin{equation}
    \mathrm{d}\theta_{l}/\mathrm{d}t \propto -\partial \mathcal{F}_t/\partial \theta_{l}|_{x=x_{l,t}^*}
\end{equation}
The operation of PCNs involves two distinct phases. Initially, the activation $x_{l,t}$ undergoes updates to minimize the free energy until they attain equilibrium. Subsequently, the weights $\theta_{l}$ undergo a single-step update based on the equilibrium values of the activations $x_{l,t}^*$. These phases are commonly referred to as inference and learning.

PC converts the feedforward pass of an ANN (artificial neural network) into an inference problem, which requires manipulating the activations of the ANN layers under certain constraints on the input or output layers, or both. The uncertainty about the optimal activations is represented by a Gaussian distribution with a mean given by the top-down prediction from the higher layer. Importantly, this inference problem is solved dynamically in each inference stage. Furthermore, the conditioning variables can be modified adaptively according to the task demands. During execution, the PCN (predictive coding network) utilizes its learned generative model embedded in the weights $\theta_t$, enabling it to cope with diverse inference problems. This demonstrates the enhanced flexibility of PCNs over ANNs, as evidenced by recent studies.\par
A common assumption is that PC provides the best inversion method for hierarchical Gaussian generative models. However, this assumption does not hold when compared to the state-of-the-art generative AI systems that employ deep neural networks, which can achieve superior performance in various domains. Moreover, there are several aspects of cortical microcircuitry remain which have so far resisted simple interpretation while using the PC framework. As such, predictive coding might be 'right' in some sense, but still missing core aspects of the computation that actually goes on in the cortex. Memory is another vital function of the brain. It seems that cortical areas implement short-term and long-term memory by means of persistent neural activity. However, modeling these memory processes within a predictive coding framework poses significant challenges.

\subsection{Forward-Forward Algorithm}
Proposed by Geoffrey Hinton, the Forward-Forward algorithm is a new learning procedure to update the weights of the network which is more biologically plausible than backpropagation. Compared with the backpropagation algorithm, the Forward-Forward algorithm updates the weights of each layer with two forward propagation processes, specifically, it updates network weights layer by layer in place, without the need to store neural activities or wait for the backpropagation of error gradients. In addition, it does not require full knowledge of the forward calculation function therefore it can tolerate black-box functions. These properties show greater biological plausibility, more akin to the functioning of the cerebral cortex. So this algorithm deserves our attention.

With the advantages of the algorithm established, it’s pertinent to explore its basic design. Generally, during the training phase, the algorithm will use two forward passes to replace the forward and backward passes, one with positive data and the other with negative data. Each layer has its own objective function, as a result, the train of each layer is independent, and the parameter updates of the former layers do not depend on the activities of the neurons in the layers behind them. The training target of each layer is to increase the goodness of positive data and decrease the goodness of negative data. In other words, the training objective for each layer is to be able to differentiate between positive data and negative data. For a given layer, The probability that an input vector is positive can be formulated as below:
\begin{equation}
    p(\text{positive}) = \sigma \left( \sum_j y_j^2 - \theta \right)
\end{equation}
where $y_j$ is the activity of hidden unit $j$,$\theta$ is a threshold, and $\sigma$ is the logistic function.

Nowadays the Forward-Forward algorithm shows comparable speed and accuracy on some small problems\cite{hinton2022forward}, but for large models, its performance is inferior to the backpropagation algorithm. Additionally, the Forward-Forward algorithm still has points that do not satisfy biological plausibility, for example, the output of each layer is still continuous signals rather than spikes, and the error signals are unsigned. In conclusion, while the Forward-Forward algorithm exhibits significant strengths and an innovative design in terms of biological plausibility, it is not without limitations. This, however, does not diminish its potential. It is the hope that this exploration of the Forward-Forward algorithm will contribute to further understanding of biological plausibility.

\subsection{Perturbation Learning}
The concept of ``perturbation learning'' was first proposed by \cite{Williams92}, which presented an associative reinforcement learning algorithm for networks containing stochastic units.
Then, \cite{WerfelXS03} introduced a perturbation learning paradigm to linear feedback neural networks.
In recent years, perturbation learning methods for training neural networks mostly mean utilizing a random perturbation on network components.
If we can observe a decrease in the error, this perturbation will be accepted.
Otherwise, we will reject this perturbation and try another perturbation.
There are generally two perturbation learning methods nowadays: weight perturbation and node perturbation.
While perturbation learning may necessitate multiple iterations for neural networks to achieve convergence, it can be considered a biologically plausible training methodology. 
This characterization arises from the absence of direct gradient descents guided by global targets in the process.

\subsubsection{Weight Perturbation}
The technique that employs perturbations to rectify the connection weights of a learning machine, such as neural networks, is referred to as weight perturbation learning.
Weight perturbation learning has been introduced as a learning rule involving the addition of perturbations to the learnable parameters of neural networks.
The generalization performance of weight perturbation learning has been scrutinized through statistical mechanical methods, revealing an asymptotic generalization property analogous to perceptron learning.

Following \cite{WerfelXS03}, we denote the input of a neural network as $x$, the learnable weights of a neural network as $w$, and then the output can be represented as $y=wx$.
Denote desired corresponding outputs is $d$, and assume there is an ideal weight $w^*$ such that $d=w^*x$.
If we choose mean square error as our loss function, then the loss can be defined as:
\begin{equation}\label{wp:mse_loss}
    L=\frac12|y-d|^2=\frac12|(w-w^*)x|^2=\frac12|Wx|^2
\end{equation}
where $W=w-w^*$.
For weight perturbation, there will be a noise matrix $\psi$ from a Gaussian distribution with mean $0$ and variance $\sigma^*$ to perturb the $W$.

\begin{equation}
    L_{\mathrm{WP}}^{\prime} = \frac12|(W+\psi)x|^2
\end{equation}

If the loss $L_{\mathrm{WP}}^{\prime}$ decreases, this perturbation $\psi$ on $W$ will be accepted.
Therefore, the weight $w$ will be updated to $w'$:

\begin{equation}
    w'=w+\Delta w = w-\frac\eta{\sigma^2}(L_{\mathrm{WP}}^{\prime}-L)\psi 
\end{equation}
where $\eta > 0$ is the learning rate.
Then repeat in this way until there are no perturbations $\psi$ that can decrease the $L_{\mathrm{WP}}^{\prime}$.

\subsubsection{Node Perturbation}
Different from weight perturbation learning, node perturbation learning constitutes a variant of the statistical gradient descent algorithm applicable to scenarios where the objective function is not explicitly defined, especially in reinforcement learning.
This method approximates the gradient of the objective function by assessing changes in the objective function resulting from perturbations.
The baseline, denoting the objective function value for an unperturbed output, plays a pivotal role in this process.
This approach can be conceptualized as reinforcement learning with a scalar reward, wherein all weight vectors are adjusted based on the scalar reward, in contrast to gradient methods that employ target vectors.
Consequently, node perturbation learning exhibits versatility, serving not only as a valuable neural network learning algorithm but also amenable to formulation as reinforcement learning or application within a brain model.

Again, we choose Equation \ref{wp:mse_loss} as the loss function.
Different from weight perturbation, node perturbation introduces a random noise matrix $\xi$ on output rather than weight:
\begin{equation}
    L_{\mathrm{NP}}^{\prime}=\frac12|Wx+\xi|^2
\end{equation}

If the loss $L_{\mathrm{NP}}^{\prime}$ decreases, this perturbation $\xi$ will be accepted.
Therefore, the weight $w$ will be updated to $w'$:
\begin{equation}
    w'=w+\Delta w = w-\frac\eta{\sigma^2}(L_{\mathrm{NP}}^{\prime}-L)\xi x^T
\end{equation}

Then repeat in this way until there are no perturbations $\xi$ that can decrease the $L_{\mathrm{NP}}^{\prime}$.

\subsubsection{Forward Gradient Learning}
{
    Forward Gradient learning is a recently raised method that implements perturbation learning using the forward-mode AD technique \cite{baydin2022gradients, silver2021learning}. In the context of biological plausibility, Forward Gradient learning dispenses with the need for weight transport and necessitates only a single-phase update in the training process. Notably, the target signal is conveyed through the feed-forward process, distinguishing it from backpropagation. Furthermore, recent studies have demonstrated the applicability of this training algorithm to Spiking Neural Networks (SNNs), emphasizing its promise in structures more akin to the human brain.
    
    Forward-mode Automatic Differentiation (AD) was first proposed by \cite{wengert1964simple}. For a given function $f$, forward-mode AD computes the matrix-vector Jacobian product $J_fv$, which is defined as the directional gradient at x with a perturbation vector $v$:
    \begin{equation}
        J_f v := \lim_{\delta \to 0} \frac{f(x + \delta v) - f(x))}{\delta}
    \end{equation}
    Note that the Jacobian product $J_fv$ is computed in a single forward evaluation. On the contrary, reverse-mode Automatic Differentiation (AD), commonly referred to as backpropagation, computes the vector Jacobian product $v^T J_f$ through a combination of forward and backward passes.  Given the efficiency of forward-mode AD in comparison to backpropagation, researchers have been exploring the implementation of learning methods utilizing forward-mode AD. The majority of deep learning approaches optimize the parameters of neural architectures through gradient descent, although calculating the exact gradient is computationally demanding in terms of both space and time. In contrast, Forward Gradient learning utilizes directional derivatives at randomly selected directions as an unbiased estimator for the true gradient, which is then used to facilitate gradient descent. Despite introducing some additional variance to the gradient estimate, this approach incurs significantly lower computational costs compared to computing the true gradient. Furthermore, the elimination of the need for back-propagation in this process enhances its biological plausibility.

    While this methodology proves adequate for addressing small-scale problems, when applied to large-scale neural architectures, the expanded dimensions within the loss space introduce a greater degree of perturbation possibilities. Consequently, this leads to increased computational expenses and diminished effectiveness. To address this issue, greedy local learning objectives have been introduced to scale this method to more intricate tasks \cite{ren2022scaling}. The same study also proposed an alternative method for estimating gradients, which involves activity perturbation instead of weight perturbation within the same framework, and notably, it yielded improved performance.
}

\subsection{Local Losses}

The local Losses method \cite{BengioLPL06,Bengio14} represents a paradigm shift from global error correction to layer-specific training.
Instead of propagating a single global error signal backward through the entire network, Local Losses employs local error signals for each layer. Each layer is equipped with its own classification layer that generates an error signal used to update that specific layer's parameters.
This approach aligns more closely with biological neural processes, where learning appears to be more distributed and locally governed.

Formally, consider a neural network composed of \(L\) layers. Let \(h^l\) denote the activations of the \(l\)-th layer, where \(h^l = f^l(h^{l-1})\) and \(f^l\) represents the transformation applied by layer \(l\). In traditional back-propagation, the overall loss \( \mathcal{L}_{\text{global}} \) depends on the final output \(h^L\) and the target labels \(y\):

\[ \mathcal{L}_{\text{global}} = \mathcal{L}(h^L, y) \]

Gradients of this global loss with respect to each layer's parameters are computed and propagated backward from the output layer to the input layer. Conversely, in the Local Losses method, each layer \(l\) has its own local loss function \( \mathcal{L}^l \), which depends on the activations \( h^l \) and a set of auxiliary target labels \( y^l \):

\[ \mathcal{L}^l = \mathcal{L}(h^l, y^l) \]

The auxiliary targets \( y^l \) can be derived from the true labels \( y \) or be unsupervised signals appropriate for the task at hand.

Each layer \(l\) is paired with a classifier \( C^l \), which maps the activations \( h^l \) to a prediction \( \hat{y}^l \):

\[ \hat{y}^l = C^l(h^l) \]

The local loss \( \mathcal{L}^l \) for layer \(l\) can be written as:

\[ \mathcal{L}^l = \mathcal{L}(C^l(h^l), y^l) \]

The parameters of both the transformation \( f^l \) and the classifier \( C^l \) are updated based on the gradient of this local loss:

\[ \theta^l \leftarrow \theta^l - \eta \frac{\partial \mathcal{L}^l}{\partial \theta^l} \]

where \( \theta^l \) represents the parameters of layer \(l\), and \( \eta \) is the learning rate.

This layer-specific training allows each layer to learn independently, guided by its own local loss. This method mimics the potentially distributed nature of learning in the brain, where different regions may adapt based on localized feedback.
The Local Losses approach offers several benefits, including biological plausibility, scalability, and robustness.
However, this approach also presents challenges:
1. Auxiliary Target Design: Designing appropriate auxiliary targets \( y^l \) for each layer can be complex and may require domain-specific knowledge;
2. Coordination: Ensuring that independently trained layers work harmoniously to achieve the overall task requires careful architectural and procedural design.


\subsection{Energy-based Learning}
{
    The continuous Hopfield models \cite{hopfield1984neurons} are recurrent neural networks that serve as content-addressable memory systems. An appropriate ``energy'' function is constructed that is always decreased by any state change produced by the activity of each neuron. The energy function's minima correspond to the preferred states of the model. The examination of iterations of Hopfield models continues to be a subject of ongoing research over several decades, including the exploration of diverse learning mechanisms. Hebbian learning and the Storkey learning rule represent two established traditional learning approaches in Hopfield models, while in recent years, a novel learning paradigm known as equilibrium propagation \cite{scellier2017equilibrium} has been introduced. 
    
    The combination of Equilibrium Propagation with Hopfield models is regarded as having significant biological plausibility. Hopfield models are commonly seen as approximations of human memory systems, deviating from the typical layered architecture observed in most artificial neural networks (ANNs). The weight update functions of Equilibrium Propagation can be constructed by synaptic learning rules based on pre- and post-synaptic activities. While this algorithm, akin to Back-propagation, calculates the gradient of the objective function during the second phase, it is notable that the neural computations in both phases are consistent, enhancing the possibility of a biological implementation. 

    Equilibrium Propagation comprises a two-phase training process. In the initial phase, a prediction is generated by fixing the inputs and allowing the Hopfield network to converge to a local energy function minimum. Subsequently, during the second phase, the outputs are adjusted toward their desired targets, and the network converges to a new state with a marginally reduced prediction error. Temporal derivatives of the neural activities in Equilibrium Propagation and Back-propagation have been proved equal by \cite{scellier2019equivalence}. This algorithm also exhibits a connection to contrastive Hebbian learning, as it learns the second phase fixed point by reducing the total energy and in comparison to the first phase fixed point achieved through prediction. In the wake of the introduction of Equilibrium Propagation, it has been extended to encompass Convolutional and Spiking Neural Networks \cite{laborieux2021scaling, o2019training}, highlighting a compelling approach as a biologically plausible method for gradient computation in deep neural networks. However, an existing issue is that Hopfield models require symmetric connections between nodes, and there is yet no biological evidence supporting the existence of a symmetric connection among actual neurons in the brain \cite{scellier2017equilibrium}. 
}

\section{Biologically Plausibility}

As depicted in Table \ref{tab:bio_analysis}, we assess the biological plausibility of the aforementioned algorithms based on the criteria outlined in Section \ref{sec:preliminary}.

\begin{table}[htp]
\small
\centering
\resizebox{\linewidth}{!}{
\begin{tabular}{l:c:c:c:c:c}
\toprule
Algorithms & Asymmetry Weights & Local Error & Non-parallel Training & Neuron Model & Unsigned Errors \\
\hline
Hebbain Learning & \cmark & \cmark & \cmark & \cmark & \cmark\\ 
STDP & \cmark & \cmark & \cmark & \cmark & \cmark\\
Feedback Alignment & \cmark & \cmark & & \\
Target Propagation & \cmark & \cmark & & \\
Predictive Coding & & \cmark &  & \cmark & \\
Foward-Forward & \cmark & \cmark & \cmark & \\
Perturbation Learning & \cmark & \cmark & \cmark & \cmark \\
Local Losses &  & \cmark & & \cmark \\
Energy-based Learning & \cmark & \cmark & \cmark &  \\
\bottomrule
\end{tabular}
}
\caption{The biological plausibility of various algorithms with the proposed criteria. Note that the criteria ``Neuron Model'' indicates whether the algorithm can be realized as a spiking neuron version.}
\label{tab:bio_analysis}
\end{table}

From Table \ref{tab:bio_analysis}, we observe that:
\begin{itemize}
    \item For Hebbian learning and STDP algorithms, they are the most biologically plausible methods, as they meet all the criteria.  Hebbian learning aligns with the principle of neurons that fire together wire together, and STDP refines this by adjusting synaptic strength based on the precise timing of spikes.  Both algorithms avoid the use of backpropagation, relying on local information for synaptic changes, and do not require symmetric weights, thereby adhering closely to biological processes.
    \item Following them, perturbation learning also exhibits biological plausibility, but it still has issues with negative values in error propagation.  Perturbation learning makes small adjustments to synaptic weights based on the perturbation of the network's output, which aligns with local error processing.  However, the presence of negative error signals does not perfectly match the typically unsigned nature of biological error signals.
    \item For forward-forward algorithm and energy-based learning, they succeed in satisfying three criteria, i.e., asymmetry weights, local error representation, and non-parallel training.  The forward-forward algorithm simplifies the learning process by overlapping the phases of forward and backward propagation, which is more in line with biological systems.  Energy-based learning focuses on minimizing a global energy function but still uses local errors and maintains weight asymmetry.  Both methods, however, do not address the spiking nature of biological neurons or the unsigned nature of biological errors.
    \item Feedback alignment and target propagation mainly focus on the learnable weight.  Therefore, they only match two criteria, i.e., asymmetry weights and local error.  Feedback alignment replaces the backpropagation of errors with a random feedback matrix, ensuring weight asymmetry and local error signals, but it still involves a form of parallel training.  Target propagation, while adjusting weights to minimize output error, also relies on weight asymmetry and local errors but does not satisfy other biological criteria.
    \item Lastly, Local Losses is the least biologically plausible method, as it is essentially the back-propagation algorithm trained layer by layer.  This method does not incorporate the asymmetry of weights, non-parallel training, or the use of local error signals.  It operates in a staged manner, unlike continuous biological learning, and does not address neuron models or unsigned error signals.  The criteria "Neuron Model" indicates whether the algorithm can be realized as a spiking neuron version, highlighting the importance of mimicking the spiking behavior of biological neurons.
\end{itemize}

\section{Experiments}

In this section, we conduct a series of comprehensive experiments.
Firstly, we empirically evaluate the performance of all biologically plausible algorithms across diverse model structures and datasets on image classification tasks.
Moreover, in order to assess how well these algorithms can present human brains' activity patterns, we innovatively compare the feature representations learned by algorithms with non-invasive brain activity records.

\subsection{Datasets}

In this section, we will introduce the datasets we utilized 

\textbf{MNIST} \quad The MNIST dataset consists of 70,000 grayscale images of handwritten digits, divided into 60,000 training images and 10,000 testing images. Each image is 28x28 pixels, and the digits range from 0 to 9. This dataset is a benchmark in the field of machine learning and image classification due to its simplicity and ease of use, providing a solid baseline for evaluating the performance of different algorithms.

\textbf{CIFAR} \quad The CIFAR dataset, created by the Canadian Institute For Advanced Research, is a widely-used benchmark in machine learning and computer vision, consisting of two versions: CIFAR-10 and CIFAR-100. CIFAR-10 contains 60,000 32x32 color images in 10 classes, split into 50,000 training and 10,000 testing images, covering categories such as airplanes, cars, and animals. CIFAR-100 extends this to 100 classes grouped into 20 superclasses, with the same train-test split ratio.

\textbf{Haxby} \quad The Haxby dataset is a neuroimaging dataset that records brain activity using functional Magnetic Resonance Imaging (fMRI) while subjects view images from different categories. This dataset is particularly valuable for our research as it allows us to compare the feature representations learned by biologically plausible algorithms with actual brain activity patterns. The dataset includes multiple subjects and a variety of visual stimuli, making it ideal for studying how well machine-learning models can mimic human brain processing.

By utilizing these datasets, we aim to provide a comprehensive evaluation of the algorithms' performance across simple and complex visual tasks, as well as their ability to replicate human brain activity patterns.

\subsection{Implementation Details}

In our experiments, we employed two primary types of models: Convolutional Neural Networks (CNNs) and Multi-Layer Perceptrons (MLPs). The CNN model consists of three convolutional layers with kernel sizes of \(5 \times 5\), \(5 \times 5\), and \(3 \times 3\), respectively, and channel sizes increasing from input to 64, 128, and 256, with a final fully connected (FC) layer mapping to the output classes. Each convolutional layer is followed by a max pooling layer with a kernel size of \(2 \times 2\). The activation function used throughout the CNN is ReLU. The MLP model is composed of an input layer, two hidden layers with 1024 and 256 units respectively, and an output layer, all utilizing ReLU activation functions. Both models were trained using a standard backpropagation algorithm with categorical cross-entropy loss and optimized with stochastic gradient descent (SGD) with momentum. Hyperparameters, such as learning rate and batch size, were fine-tuned based on validation set performance. This implementation enabled us to evaluate and compare the performance of biologically plausible algorithms across different network architectures and datasets, providing insights into their effectiveness in mimicking human brain activity patterns.

Specifically, for the STDP method, we utilize the Pytorch-based framework called SpikingJelly \cite{fang2023spikingjelly} which is tailored for spiking neural networks.

\subsection{Image Classification}

We evaluate all bio-plausible algorithms on image classification benchmarks, reported in Table \ref{tab:main_table}.

\begin{table}[htp]
\centering
\resizebox{\linewidth}{!}{
\begin{tabular}{l:cc:cc:cc}
\toprule
\hline
\multirow{2}{*}{\bf Algorithm} & \multicolumn{2}{c:}{\bf MNIST} & \multicolumn{2}{c:}{\bf CIFAR-10} & \multicolumn{2}{c}{\bf CIFAR-100} \\
& MLP & CNN & MLP & CNN & MLP & CNN\\
\hline
Backpropagation & $\bf98.62${\scriptsize $\pm 0.17\%$} & $\bf99.59${\scriptsize $\pm 0.09\%$} & $\bf58.60${\small $\pm 0.22\%$} & $\bf73.64${\scriptsize $\pm 0.23\%$} &  $\bf34.76${\scriptsize $\pm 0.20\%$} &  $\bf52.45${\scriptsize $\pm 0.28\%$} \\ \hline
Hebbian Learning & $78.29${\scriptsize $\pm 0.07\%$} & $83.05${\scriptsize $\pm 0.12\%$} & $19.98${\small $\pm 0.23\%$} & $29.86${\scriptsize $\pm 0.13\%$} & $-$ & $-$ \\
STDP & $77.18${\scriptsize $\pm 0.17\%$} & $91.67${\scriptsize $\pm 0.04\%$} & $22.68${\scriptsize $\pm 0.3\%$} & $33.19${\scriptsize $\pm 0.38\%$} 
& $-$ & $-$ \\
Feedback Alignment & $91.87${\scriptsize $\pm 0.08\%$} & $97.00${\scriptsize $\pm 0.13\%$} & $48.46${\scriptsize $\pm 0.11\%$} & \underline{$59.60$}{\scriptsize $\pm 0.46\%$} & \underline{$20.75$}{\scriptsize $\pm 0.18\%$} & \underline{$33.30$}{\scriptsize $\pm 0.20\%$} \\
Direct Feedback Alignment & \underline{$97.47$}{\scriptsize $\pm 0.15\%$} & $97.69${\scriptsize $\pm 0.14\%$} & \underline{$49.85$}{\scriptsize $\pm 0.21\%$} & $58.38${\scriptsize $\pm 0.39\%$} & $17.17${\scriptsize $\pm 0.16\%$} & $22.91${\scriptsize $\pm 0.50\%$}\\
Target Propagation & $94.01${\scriptsize $\pm 0.12\%$} & $96.40${\scriptsize $\pm 0.05\%$} & $46.10${\scriptsize $\pm 0.10\%$} & $50.88${\scriptsize $\pm 0.07\%$} & $15.70${\scriptsize $\pm 0.22\%$} & $21.20${\scriptsize $\pm 0.38\%$}\\
Predictive coding & $97.41${\scriptsize $\pm 0.14\%$} & \underline{$99.12$}{\scriptsize $\pm 0.10\%$} & $46.53${\scriptsize $\pm 0.23\%$} & $56.12${\scriptsize $\pm 0.44\%$} & $10.32${\scriptsize $\pm 0.05\%$} & $16.32${\scriptsize $\pm 0.12\%$}\\ 
Forward-Forward & $96.99${\scriptsize $\pm 0.14\%$} & $15.66${\scriptsize $\pm 0.08\%$} & $39.48${\scriptsize $\pm 0.10\%$} & $-$ & $-$ & $-$ \\
Perturbation Learning & $91.44${\scriptsize $\pm 0.43\%$} & $92.61${\scriptsize $\pm 0.43\%$} & $31.07${\scriptsize $\pm 0.31\%$} & $39.72${\scriptsize $\pm 0.38\%$} & $-$ & $-$ \\
Local Losses & $\bf 98.56${\scriptsize $\pm 0.19\%$} & $\bf 99.39${\scriptsize $\pm 0.06\%$} & $\bf 55.18${\scriptsize $\pm 0.13\%$} & $\bf 72.18${\scriptsize $\pm 0.10\%$} & $\bf 28.83${\scriptsize $\pm 0.21\%$} & $\bf 35.39${\scriptsize $\pm 0.14\%$} \\
Equilibrium Propagation & $93.81${\scriptsize $\pm 0.18\%$} & $26.72${\scriptsize $\pm 0.22\%$} & $16.93${\scriptsize $\pm 0.10\%$} & $-$ & $-$ & $-$ \\
\hline
\bottomrule
\end{tabular}
}
\caption{The accuracy of image classification achieved by various bio-plausible learning algorithms.
The best and the second-placed results of bio-plausible algorithms are formatted in bold font and underlined format.
Note that $-$ denotes that the models \textbf{fail to converge} when training under this setting.
All results are averaged across at least $3$ random seeds.
}
\label{tab:main_table}
\end{table}

As shown in Table \ref{tab:main_table}, we can draw conclusion that:
\textbf{(1)} Local losses, being the most similar to backpropagation, achieve performance closest to traditional backpropagation. This aligns with our expectations, as layer-by-layer training of neural networks still relies on global label signals, utilizing local errors exclusively for weight updates.
\textbf{(2)} While Hebbian learning and STDP are the most biologically plausible algorithms, they exhibit significantly poorer performance compared to alternative methods. Particularly, they both fail to converge on the CIFAR-100 dataset.
\textbf{(3)} The forward-forward algorithm and equilibrium propagation (an energy-based method) appear effective primarily with the MLP architecture on tasks such as MNIST. However, these methods encounter challenges and fail to converge when applied to more complex tasks or architectures like CNNs.

Taking Table \ref{tab:bio_analysis} into consideration, it is evident that \textbf{the current bio-plausible learning algorithms, while showing promise, often fall short of achieving the high performance demonstrated by traditional backpropagation}, especially on more complex datasets like CIFAR-10 and CIFAR-100. Algorithms like Feedback Alignment, Direct Feedback Alignment, Predictive Coding, and Local Losses demonstrate competitive performance on simpler datasets such as MNIST, indicating their potential viability. However, their diminished effectiveness on more challenging tasks underscores the need for further advancements in this field.

For the community to make significant strides, \textbf{it is necessary to develop a learning algorithm that seamlessly integrates biological plausibility with high performance}. Such an algorithm would need to address the key criteria of biological learning, including asymmetry of forward and backward weights, local error representation, non-parallel training, realistic neuron models, and unsigned error signals, without compromising on accuracy and scalability. This endeavor will likely involve innovative approaches that combine insights from neuroscience with advanced machine learning techniques.

The development of such an algorithm would not only bridge the gap between biological realism and computational efficiency but also potentially lead to more robust, efficient, and adaptable learning systems. It would pave the way for a new era of machine learning that is not only inspired by but also closely aligned with the principles of biological learning, ultimately enhancing our ability to develop intelligent systems that can learn and adapt in more human-like ways.

\subsection{Comparative Analysis of Representations Learned by Algorithms and Activity Patterns Elicited by Human Brains}

To investigate the biological plausibility of the previously discussed algorithms, we have developed a method to compute representation similarity between biologically plausible algorithms and the human brain.
Specifically, we utilize NeuroRA, a toolbox designed for representation similarity analysis.
NeuroRA allows us to acquire various types of neural data, such as fMRI and ROI, and to compute the Representational Dissimilarity Matrix (RDM).
Using the Haxby dataset, which is compatible with NeuroRA, we train our model and compute an average representation for each category within the dataset.
Based on these average representations, we calculate the RDM for our model.
Subsequently, we measure the similarity between the RDMs of human brains and our model using cosine similarity, which serves as our similarity metric.
The results are presented in Table \ref{tab:similarity}.

\begin{table}[htp]
\centering
\setlength{\tabcolsep}{15pt}
\small

\begin{tabular}{l:c:cc}
\toprule
\hline
\multirow{2}{*}{\bf Algorithm} & \multirow{2}{*}{\bf Metric} & \multicolumn{2}{c}{\bf Haxby} \\
& & MLP & CNN\\
\hline
\multirow{2}{*}{Backpropagation} & Accuracy$\uparrow$ & $\bf67.16\%$ & $\bf79.10\%$ \\ 
& Similarity$\uparrow$ & $0.799$ & $0.795$ \\ \hline
\multirow{2}{*}{Feedback Alignment} & Accuracy$\uparrow$ & $58.03\%$ & $63.55\%$ \\ 
& Similarity$\uparrow$ & $0.796$ & $0.793$ \\ \hline
\multirow{2}{*}{Predictive coding} & Accuracy$\uparrow$ & $57.86\%$ & $64.70\%$ \\ 
& Similarity$\uparrow$ & $0.805$ & $\bf0.806$ \\ \hline
\multirow{2}{*}{Local Losses} & Accuracy$\uparrow$ & $64.08\%$ & $69.85\%$ \\ 
& Similarity$\uparrow$ & $0.794$ & $0.798$ \\ \hline
\multirow{2}{*}{Forward-Forward} & Accuracy$\uparrow$ & $55.22\%$ & $--$ \\ 
& Similarity$\uparrow$ & $\bf0.813$ & $--$ \\ 
\hline
\bottomrule
\end{tabular}
\caption{
Table of classification accuracy and RDM similarity of backpropagation and $4$ selected bio-plausible algorithms.
Note that we chose $4$ algorithms because they were the only ones that successfully converged during training.
For CNN trained by the forward-forward algorithm, it fails to converge.
$\uparrow$ indicates the higher the better.
The best results are formatted in bold font.
}
\label{tab:similarity}
\end{table}

Note that we chose $4$ algorithms because they were the only ones that successfully converged during training. For CNN trained by the forward-forward algorithm, it fails to converge.
From Table \ref{tab:similarity}, we find that:
\textbf{(1)} Backpropagation significantly outperforms other bio-plausible algorithms on accuracy. This observation is reasonable and aligns with findings in datasets like MNIST and CIFAR.
\textbf{(2)} For RDM similarity, Backpropagation is not the top performer; predictive coding shows the highest similarity. This suggests that although backpropagation achieves high accuracy, it might not be as biologically plausible in terms of the representation it learns.
\textbf{(3)} There appears to be a trade-off between accuracy and similarity.  Backpropagation achieves relatively high RDM similarity alongside its superior performance, suggesting that high accuracy might contribute to higher similarity.  This leads us to speculate that if the performance of algorithms like predictive coding and forward-forward were improved to match the accuracy of backpropagation, they might exhibit even higher similarity scores.  This hypothesis indicates that enhancing the accuracy of biologically plausible algorithms could potentially increase their alignment with human neural representations, offering a promising direction for future research in developing models that are both effective and biologically plausible.

\section{Open Research Questions}

The field of biologically plausible deep learning is in its nascent stages, and there are several \textbf{critical research questions} that need to be addressed to advance the state-of-the-art.
This discussion aims to delve into these questions, exploring potential pathways and challenges associated with scaling these models, learning temporal sequences, and optimizing neural circuit dynamics.

\subsection{Scaling Biologically Plausible Implementations}
Biologically inspired models often face challenges in scalability due to their intricate architectures and the need to adhere to the constraints of biological plausibility.  Traditional deep learning models have demonstrated remarkable success in handling complex, high-dimensional tasks, such as image and speech recognition, but they often rely on computational techniques and resources that are not biologically feasible.  To bridge this gap, researchers must explore methods to enhance the scalability of biologically plausible models without compromising their inherent principles.  This involves investigating new algorithms, optimizing hardware implementations, and perhaps most critically, developing hybrid approaches that combine the strengths of both biologically inspired and traditional deep learning methods.

\subsection{Learning Temporal Sequences}
Biological networks excel at processing temporal sequences, a capability that remains a significant challenge for artificial systems.  The dynamic and recurrent nature of biological neurons allows for sophisticated temporal processing, enabling organisms to navigate, predict, and learn from sequential events in their environments.  In contrast, most artificial neural networks, particularly those used in deep learning, struggle with temporal dependencies unless specifically designed with recurrent or attention-based mechanisms.  Developing biologically plausible methods for learning temporal sequences could involve leveraging the properties of spiking neural networks (SNNs) or other neuromorphic computing approaches that naturally accommodate time-dependent processing.  Additionally, understanding how biological networks balance short-term and long-term memory and applying these principles to artificial systems could lead to more robust and efficient temporal learning models.

\subsection{Optimizing Neural Circuit Dynamics}
Biological systems are not only efficient in learning but also in their energy consumption and adaptive capabilities.  The optimization of neural circuit dynamics to support efficient learning is a multifaceted problem that encompasses the minimization of energy use, the maximization of learning speed, and the enhancement of adaptability to changing environments.  Biological neurons exhibit a variety of dynamic behaviors, such as synaptic plasticity and homeostasis, which contribute to their learning efficiency.  Translating these dynamics into artificial systems requires a deep understanding of the underlying biological processes and the development of algorithms that can mimic these processes effectively.  This might involve creating more sophisticated models of synaptic plasticity, incorporating adaptive mechanisms that allow artificial networks to self-tune in response to environmental changes, and designing hardware that can support these dynamic processes efficiently.

\subsection{Fully Biologically Plausible Algorithms}
The development of fully biologically plausible algorithms is a crucial area of research within the field of biologically inspired deep learning. These algorithms aim to replicate the principles and mechanisms of learning observed in biological systems, such as local learning rules, Hebbian plasticity, and the balance between excitatory and inhibitory signals. Achieving full biological plausibility involves not only mimicking the structural and functional aspects of biological neurons but also adhering to the constraints of biological systems, such as limited energy resources and real-time processing capabilities.

\subsection{Hardware Implementation}
The advancement of hardware specifically designed for biologically plausible models is vital for the practical application of these systems. Neuromorphic chips, such as those designed for spiking neural networks (SNNs) \cite{Maas1997NetworksOS}, represent a significant step towards this goal. These chips, like IBM's TrueNorth \cite{akopyan2015truenorth} and Intel's Loihi \cite{davies2018loihi}, aim to emulate the structure and functionality of biological neural circuits, enabling more efficient processing of information through event-based computation and asynchronous communication. Unlike traditional processors, neuromorphic hardware can inherently handle the temporal dynamics and adaptive behaviors characteristic of biological systems. This includes the ability to support synaptic plasticity, dynamic learning, and low-power operation. Developing and optimizing such hardware involves overcoming challenges related to scalability, integration with existing computing infrastructure, and ensuring that the hardware can support the complex dynamics required for advanced learning tasks. By focusing on these areas, researchers can create more effective and efficient hardware solutions that bring biologically plausible deep learning closer to practical reality.






\section{Conclusion}

In this study, we first established fix criteria for biological plausibility and applied them to assess a range of existing representative algorithms, including Hebbian learning, spike-timing-dependent plasticity, feedback alignment, target propagation, predictive coding, forward-forward algorithm, perturbation learning, local losses, and energy-based learning, across diverse network architectures and datasets.  
Additionally, we compared the feature representations learned by these algorithms with non-invasive brain activity records under identical stimuli to identify those that most accurately replicate brain activity patterns.
We find that predictive coding and forward-forward algorithm achieve the largest similarity with brain activity patterns, indicating these two methods not only show considerable performance but also have a certain biological plausibility.
Our findings provide a comprehensive assessment and insights into biologically plausible algorithms, aiming to inspire new developments that bridge the gap between neuroscience and machine learning.

\bibliography{main}


\begin{thebibliography}{78}
\ifx \bisbn   \undefined \def \bisbn  #1{ISBN #1}\fi
\ifx \binits  \undefined \def \binits#1{#1}\fi
\ifx \bauthor  \undefined \def \bauthor#1{#1}\fi
\ifx \batitle  \undefined \def \batitle#1{#1}\fi
\ifx \bjtitle  \undefined \def \bjtitle#1{#1}\fi
\ifx \bvolume  \undefined \def \bvolume#1{\textbf{#1}}\fi
\ifx \byear  \undefined \def \byear#1{#1}\fi
\ifx \bissue  \undefined \def \bissue#1{#1}\fi
\ifx \bfpage  \undefined \def \bfpage#1{#1}\fi
\ifx \blpage  \undefined \def \blpage #1{#1}\fi
\ifx \burl  \undefined \def \burl#1{\textsf{#1}}\fi
\ifx \doiurl  \undefined \def \doiurl#1{\url{https://doi.org/#1}}\fi
\ifx \betal  \undefined \def \betal{\textit{et al.}}\fi
\ifx \binstitute  \undefined \def \binstitute#1{#1}\fi
\ifx \binstitutionaled  \undefined \def \binstitutionaled#1{#1}\fi
\ifx \bctitle  \undefined \def \bctitle#1{#1}\fi
\ifx \beditor  \undefined \def \beditor#1{#1}\fi
\ifx \bpublisher  \undefined \def \bpublisher#1{#1}\fi
\ifx \bbtitle  \undefined \def \bbtitle#1{#1}\fi
\ifx \bedition  \undefined \def \bedition#1{#1}\fi
\ifx \bseriesno  \undefined \def \bseriesno#1{#1}\fi
\ifx \blocation  \undefined \def \blocation#1{#1}\fi
\ifx \bsertitle  \undefined \def \bsertitle#1{#1}\fi
\ifx \bsnm \undefined \def \bsnm#1{#1}\fi
\ifx \bsuffix \undefined \def \bsuffix#1{#1}\fi
\ifx \bparticle \undefined \def \bparticle#1{#1}\fi
\ifx \barticle \undefined \def \barticle#1{#1}\fi
\bibcommenthead
\ifx \bconfdate \undefined \def \bconfdate #1{#1}\fi
\ifx \botherref \undefined \def \botherref #1{#1}\fi
\ifx \url \undefined \def \url#1{\textsf{#1}}\fi
\ifx \bchapter \undefined \def \bchapter#1{#1}\fi
\ifx \bbook \undefined \def \bbook#1{#1}\fi
\ifx \bcomment \undefined \def \bcomment#1{#1}\fi
\ifx \oauthor \undefined \def \oauthor#1{#1}\fi
\ifx \citeauthoryear \undefined \def \citeauthoryear#1{#1}\fi
\ifx \endbibitem  \undefined \def \endbibitem {}\fi
\ifx \bconflocation  \undefined \def \bconflocation#1{#1}\fi
\ifx \arxivurl  \undefined \def \arxivurl#1{\textsf{#1}}\fi
\csname PreBibitemsHook\endcsname

\bibitem[\protect\citeauthoryear{Rumelhart et~al.}{1986}]{rumelhart1986learning}
\begin{barticle}
\bauthor{\bsnm{Rumelhart}, \binits{D.E.}},
\bauthor{\bsnm{Hinton}, \binits{G.E.}},
\bauthor{\bsnm{Williams}, \binits{R.J.}}:
\batitle{Learning internal representations by error propagation, parallel distributed processing, explorations in the microstructure of cognition, ed. de rumelhart and j. mcclelland. vol. 1. 1986}.
\bjtitle{Biometrika}
\bvolume{71},
\bfpage{599}--\blpage{607}
(\byear{1986})
\end{barticle}
\endbibitem

\bibitem[\protect\citeauthoryear{LeCun et~al.}{2015}]{lecun2015deep}
\begin{barticle}
\bauthor{\bsnm{LeCun}, \binits{Y.}},
\bauthor{\bsnm{Bengio}, \binits{Y.}},
\bauthor{\bsnm{Hinton}, \binits{G.}}:
\batitle{Deep learning}.
\bjtitle{nature}
\bvolume{521}(\bissue{7553}),
\bfpage{436}--\blpage{444}
(\byear{2015})
\end{barticle}
\endbibitem

\bibitem[\protect\citeauthoryear{Delorme et~al.}{2001}]{Delorme2001NetworksOI}
\begin{barticle}
\bauthor{\bsnm{Delorme}, \binits{A.}},
\bauthor{\bsnm{Perrinet}, \binits{L.U.}},
\bauthor{\bsnm{Thorpe}, \binits{S.J.}}:
\batitle{Networks of integrate-and-fire neurons using rank order coding b: Spike timing dependent plasticity and emergence of orientation selectivity}.
\bjtitle{Neurocomputing}
\bvolume{38-40},
\bfpage{539}--\blpage{545}
(\byear{2001})
\end{barticle}
\endbibitem

\bibitem[\protect\citeauthoryear{Devlin et~al.}{2019}]{Devlin2019BERTPO}
\begin{bchapter}
\bauthor{\bsnm{Devlin}, \binits{J.}},
\bauthor{\bsnm{Chang}, \binits{M.-W.}},
\bauthor{\bsnm{Lee}, \binits{K.}},
\bauthor{\bsnm{Toutanova}, \binits{K.}}:
\bctitle{Bert: Pre-training of deep bidirectional transformers for language understanding}.
In: \bbtitle{North American Chapter of the Association for Computational Linguistics}
(\byear{2019})
\end{bchapter}
\endbibitem

\bibitem[\protect\citeauthoryear{Brown et~al.}{2020}]{brown2020language}
\begin{barticle}
\bauthor{\bsnm{Brown}, \binits{T.}},
\bauthor{\bsnm{Mann}, \binits{B.}},
\bauthor{\bsnm{Ryder}, \binits{N.}},
\bauthor{\bsnm{Subbiah}, \binits{M.}},
\bauthor{\bsnm{Kaplan}, \binits{J.D.}},
\bauthor{\bsnm{Dhariwal}, \binits{P.}},
\bauthor{\bsnm{Neelakantan}, \binits{A.}},
\bauthor{\bsnm{Shyam}, \binits{P.}},
\bauthor{\bsnm{Sastry}, \binits{G.}},
\bauthor{\bsnm{Askell}, \binits{A.}}, \betal:
\batitle{Language models are few-shot learners}.
\bjtitle{Advances in neural information processing systems}
\bvolume{33},
\bfpage{1877}--\blpage{1901}
(\byear{2020})
\end{barticle}
\endbibitem

\bibitem[\protect\citeauthoryear{Bianchini et~al.}{1997}]{bianchini1997terminal}
\begin{barticle}
\bauthor{\bsnm{Bianchini}, \binits{M.}},
\bauthor{\bsnm{Fanelli}, \binits{S.}},
\bauthor{\bsnm{Gori}, \binits{M.}},
\bauthor{\bsnm{Maggini}, \binits{M.}}:
\batitle{Terminal attractor algorithms: A critical analysis}.
\bjtitle{Neurocomputing}
\bvolume{15}(\bissue{1}),
\bfpage{3}--\blpage{13}
(\byear{1997})
\end{barticle}
\endbibitem

\bibitem[\protect\citeauthoryear{Payeur et~al.}{2021}]{payeur2021burst}
\begin{barticle}
\bauthor{\bsnm{Payeur}, \binits{A.}},
\bauthor{\bsnm{Guerguiev}, \binits{J.}},
\bauthor{\bsnm{Zenke}, \binits{F.}},
\bauthor{\bsnm{Richards}, \binits{B.A.}},
\bauthor{\bsnm{Naud}, \binits{R.}}:
\batitle{Burst-dependent synaptic plasticity can coordinate learning in hierarchical circuits}.
\bjtitle{Nature neuroscience}
\bvolume{24}(\bissue{7}),
\bfpage{1010}--\blpage{1019}
(\byear{2021})
\end{barticle}
\endbibitem

\bibitem[\protect\citeauthoryear{Zahid et~al.}{2023}]{zahid2023predictive}
\begin{barticle}
\bauthor{\bsnm{Zahid}, \binits{U.}},
\bauthor{\bsnm{Guo}, \binits{Q.}},
\bauthor{\bsnm{Fountas}, \binits{Z.}}:
\batitle{Predictive coding as a neuromorphic alternative to backpropagation: A critical evaluation}.
\bjtitle{Neural Computation}
\bvolume{35}(\bissue{12}),
\bfpage{1881}--\blpage{1909}
(\byear{2023})
\end{barticle}
\endbibitem

\bibitem[\protect\citeauthoryear{Stork}{1989}]{stork1989backpropagation}
\begin{bchapter}
\bauthor{\bsnm{Stork}}:
\bctitle{Is backpropagation biologically plausible?}
In: \bbtitle{International 1989 Joint Conference on Neural Networks},
pp. \bfpage{241}--\blpage{246}
(\byear{1989}).
\bcomment{IEEE}
\end{bchapter}
\endbibitem

\bibitem[\protect\citeauthoryear{Crick}{1989}]{crick1989recent}
\begin{barticle}
\bauthor{\bsnm{Crick}, \binits{F.}}:
\batitle{The recent excitement about neural networks.}
\bjtitle{Nature}
\bvolume{337}(\bissue{6203}),
\bfpage{129}--\blpage{132}
(\byear{1989})
\end{barticle}
\endbibitem

\bibitem[\protect\citeauthoryear{Lillicrap et~al.}{2020}]{lillicrap2020backpropagation}
\begin{barticle}
\bauthor{\bsnm{Lillicrap}, \binits{T.P.}},
\bauthor{\bsnm{Santoro}, \binits{A.}},
\bauthor{\bsnm{Marris}, \binits{L.}},
\bauthor{\bsnm{Akerman}, \binits{C.J.}},
\bauthor{\bsnm{Hinton}, \binits{G.}}:
\batitle{Backpropagation and the brain}.
\bjtitle{Nature Reviews Neuroscience}
\bvolume{21}(\bissue{6}),
\bfpage{335}--\blpage{346}
(\byear{2020})
\end{barticle}
\endbibitem

\bibitem[\protect\citeauthoryear{Munakata and Pfaffly}{2004}]{munakata2004hebbian}
\begin{barticle}
\bauthor{\bsnm{Munakata}, \binits{Y.}},
\bauthor{\bsnm{Pfaffly}, \binits{J.}}:
\batitle{Hebbian learning and development}.
\bjtitle{Developmental science}
\bvolume{7}(\bissue{2}),
\bfpage{141}--\blpage{148}
(\byear{2004})
\end{barticle}
\endbibitem

\bibitem[\protect\citeauthoryear{Song et~al.}{2000}]{Song2000CompetitiveHL}
\begin{barticle}
\bauthor{\bsnm{Song}, \binits{S.}},
\bauthor{\bsnm{Miller}, \binits{K.D.}},
\bauthor{\bsnm{Abbott}, \binits{L.F.}}:
\batitle{Competitive hebbian learning through spike-timing-dependent synaptic plasticity}.
\bjtitle{Nature Neuroscience}
\bvolume{3},
\bfpage{919}--\blpage{926}
(\byear{2000})
\end{barticle}
\endbibitem

\bibitem[\protect\citeauthoryear{Lillicrap et~al.}{2016}]{Lillicrap2016RandomSF}
\begin{botherref}
\oauthor{\bsnm{Lillicrap}, \binits{T.P.}},
\oauthor{\bsnm{Cownden}, \binits{D.}},
\oauthor{\bsnm{Tweed}, \binits{D.B.}},
\oauthor{\bsnm{Akerman}, \binits{C.J.}}:
Random synaptic feedback weights support error backpropagation for deep learning.
Nature Communications
\textbf{7}
(2016)
\end{botherref}
\endbibitem

\bibitem[\protect\citeauthoryear{Bengio}{2014}]{Bengio14}
\begin{botherref}
\oauthor{\bsnm{Bengio}, \binits{Y.}}:
How auto-encoders could provide credit assignment in deep networks via target propagation.
CoRR
\textbf{abs/1407.7906}
(2014)
\end{botherref}
\endbibitem

\bibitem[\protect\citeauthoryear{Rao and Ballard}{1999}]{rao1999predictive}
\begin{barticle}
\bauthor{\bsnm{Rao}, \binits{R.P.}},
\bauthor{\bsnm{Ballard}, \binits{D.H.}}:
\batitle{Predictive coding in the visual cortex: a functional interpretation of some extra-classical receptive-field effects}.
\bjtitle{Nature neuroscience}
\bvolume{2}(\bissue{1}),
\bfpage{79}--\blpage{87}
(\byear{1999})
\end{barticle}
\endbibitem

\bibitem[\protect\citeauthoryear{Hinton}{2022}]{hinton2022forward}
\begin{botherref}
\oauthor{\bsnm{Hinton}, \binits{G.}}:
The forward-forward algorithm: Some preliminary investigations.
arXiv preprint arXiv:2212.13345
(2022)
\end{botherref}
\endbibitem

\bibitem[\protect\citeauthoryear{Williams}{1992}]{Williams92}
\begin{barticle}
\bauthor{\bsnm{Williams}, \binits{R.J.}}:
\batitle{Simple statistical gradient-following algorithms for connectionist reinforcement learning}.
\bjtitle{Machine Learning}
\bvolume{8},
\bfpage{229}--\blpage{256}
(\byear{1992})
\end{barticle}
\endbibitem

\bibitem[\protect\citeauthoryear{Werfel et~al.}{2003}]{WerfelXS03}
\begin{bchapter}
\bauthor{\bsnm{Werfel}, \binits{J.}},
\bauthor{\bsnm{Xie}, \binits{X.}},
\bauthor{\bsnm{Seung}, \binits{H.S.}}:
\bctitle{Learning curves for stochastic gradient descent in linear feedforward networks}.
In: \bbtitle{Advances in Neural Information Processing Systems},
pp. \bfpage{1197}--\blpage{1204}
(\byear{2003})
\end{bchapter}
\endbibitem

\bibitem[\protect\citeauthoryear{Marblestone et~al.}{2016}]{MarblestoneWK16}
\begin{barticle}
\bauthor{\bsnm{Marblestone}, \binits{A.H.}},
\bauthor{\bsnm{Wayne}, \binits{G.}},
\bauthor{\bsnm{K{\"{o}}rding}, \binits{K.P.}}:
\batitle{Toward an integration of deep learning and neuroscience}.
\bjtitle{Frontiers Comput. Neurosci.}
\bvolume{10},
\bfpage{94}
(\byear{2016})
\end{barticle}
\endbibitem

\bibitem[\protect\citeauthoryear{Hopfield}{1984}]{hopfield1984neurons}
\begin{barticle}
\bauthor{\bsnm{Hopfield}, \binits{J.J.}}:
\batitle{Neurons with graded response have collective computational properties like those of two-state neurons.}
\bjtitle{Proceedings of the national academy of sciences}
\bvolume{81}(\bissue{10}),
\bfpage{3088}--\blpage{3092}
(\byear{1984})
\end{barticle}
\endbibitem

\bibitem[\protect\citeauthoryear{Scellier and Bengio}{2017}]{scellier2017equilibrium}
\begin{barticle}
\bauthor{\bsnm{Scellier}, \binits{B.}},
\bauthor{\bsnm{Bengio}, \binits{Y.}}:
\batitle{Equilibrium propagation: Bridging the gap between energy-based models and backpropagation}.
\bjtitle{Frontiers in computational neuroscience}
\bvolume{11},
\bfpage{24}
(\byear{2017})
\end{barticle}
\endbibitem

\bibitem[\protect\citeauthoryear{Hebb}{1949}]{hebb1949organisation}
\begin{botherref}
\oauthor{\bsnm{Hebb}, \binits{D.O.}}:
The organisation of behaviour: a neuropsychological theory
(1949)
\end{botherref}
\endbibitem

\bibitem[\protect\citeauthoryear{Löwel and Singer}{1992}]{doi:10.1126/science.1372754}
\begin{barticle}
\bauthor{\bsnm{Löwel}, \binits{S.}},
\bauthor{\bsnm{Singer}, \binits{W.}}:
\batitle{Selection of intrinsic horizontal connections in the visual cortex by correlated neuronal activity}.
\bjtitle{Science}
\bvolume{255}(\bissue{5041}),
\bfpage{209}--\blpage{212}
(\byear{1992})
\doiurl{10.1126/science.1372754}
{\href{https://arxiv.org/abs/https://www.science.org/doi/pdf/10.1126/science.1372754}{{https://www.science.org/doi/pdf/10.1126/science.1372754}}}
\end{barticle}
\endbibitem

\bibitem[\protect\citeauthoryear{Gupta et~al.}{2021}]{gupta2021hebbnet}
\begin{bchapter}
\bauthor{\bsnm{Gupta}, \binits{M.}},
\bauthor{\bsnm{Ambikapathi}, \binits{A.}},
\bauthor{\bsnm{Ramasamy}, \binits{S.}}:
\bctitle{Hebbnet: A simplified hebbian learning framework to do biologically plausible learning}.
In: \bbtitle{ICASSP 2021-2021 IEEE International Conference on Acoustics, Speech and Signal Processing (ICASSP)},
pp. \bfpage{3115}--\blpage{3119}
(\byear{2021}).
\bcomment{IEEE}
\end{bchapter}
\endbibitem

\bibitem[\protect\citeauthoryear{Moraitis et~al.}{2022}]{moraitis2022softhebb}
\begin{barticle}
\bauthor{\bsnm{Moraitis}, \binits{T.}},
\bauthor{\bsnm{Toichkin}, \binits{D.}},
\bauthor{\bsnm{Journ{\'e}}, \binits{A.}},
\bauthor{\bsnm{Chua}, \binits{Y.}},
\bauthor{\bsnm{Guo}, \binits{Q.}}:
\batitle{Softhebb: Bayesian inference in unsupervised hebbian soft winner-take-all networks}.
\bjtitle{Neuromorphic Computing and Engineering}
\bvolume{2}(\bissue{4}),
\bfpage{044017}
(\byear{2022})
\end{barticle}
\endbibitem

\bibitem[\protect\citeauthoryear{Artola and Singer}{1993}]{artola1993long}
\begin{barticle}
\bauthor{\bsnm{Artola}, \binits{A.}},
\bauthor{\bsnm{Singer}, \binits{W.}}:
\batitle{Long-term depression of excitatory synaptic transmission and its relationship to long-term potentiation}.
\bjtitle{Trends in neurosciences}
\bvolume{16}(\bissue{11}),
\bfpage{480}--\blpage{487}
(\byear{1993})
\end{barticle}
\endbibitem

\bibitem[\protect\citeauthoryear{Lisman}{1989}]{lisman1989mechanism}
\begin{barticle}
\bauthor{\bsnm{Lisman}, \binits{J.}}:
\batitle{A mechanism for the hebb and the anti-hebb processes underlying learning and memory.}
\bjtitle{Proceedings of the National Academy of Sciences}
\bvolume{86}(\bissue{23}),
\bfpage{9574}--\blpage{9578}
(\byear{1989})
\end{barticle}
\endbibitem

\bibitem[\protect\citeauthoryear{Caporale and Dan}{2008}]{caporale2008spike}
\begin{barticle}
\bauthor{\bsnm{Caporale}, \binits{N.}},
\bauthor{\bsnm{Dan}, \binits{Y.}}:
\batitle{Spike timing--dependent plasticity: a hebbian learning rule}.
\bjtitle{Annu. Rev. Neurosci.}
\bvolume{31},
\bfpage{25}--\blpage{46}
(\byear{2008})
\end{barticle}
\endbibitem

\bibitem[\protect\citeauthoryear{Kampa et~al.}{2004}]{Kampa_Clements_Jonas_Stuart_2004}
\begin{barticle}
\bauthor{\bsnm{Kampa}, \binits{B.M.}},
\bauthor{\bsnm{Clements}, \binits{J.}},
\bauthor{\bsnm{Jonas}, \binits{P.}},
\bauthor{\bsnm{Stuart}, \binits{G.J.}}:
\batitle{Kinetics of mg2+unblock of nmda receptors: implications for spike-timing dependent synaptic plasticity}.
\bjtitle{The Journal of Physiology}
\bvolume{556}(\bissue{2}),
\bfpage{337}--\blpage{345}
(\byear{2004})
\doiurl{10.1113/jphysiol.2003.058842}
\end{barticle}
\endbibitem

\bibitem[\protect\citeauthoryear{Watanabe et~al.}{2002}]{Watanabe_Hoffman_Migliore_Johnston_2002}
\begin{barticle}
\bauthor{\bsnm{Watanabe}, \binits{S.}},
\bauthor{\bsnm{Hoffman}, \binits{D.A.}},
\bauthor{\bsnm{Migliore}, \binits{M.}},
\bauthor{\bsnm{Johnston}, \binits{D.}}:
\batitle{Dendritic k + channels contribute to spike-timing dependent long-term potentiation in hippocampal pyramidal neurons}.
\bjtitle{Proceedings of the National Academy of Sciences}
\bvolume{99}(\bissue{12}),
\bfpage{8366}--\blpage{8371}
(\byear{2002})
\doiurl{10.1073/pnas.122210599}
\end{barticle}
\endbibitem

\bibitem[\protect\citeauthoryear{Hoffman et~al.}{1997}]{Hoffman_Magee_Colbert_Johnston_1997}
\begin{barticle}
\bauthor{\bsnm{Hoffman}, \binits{D.A.}},
\bauthor{\bsnm{Magee}, \binits{J.C.}},
\bauthor{\bsnm{Colbert}, \binits{C.M.}},
\bauthor{\bsnm{Johnston}, \binits{D.}}:
\batitle{K+ channel regulation of signal propagation in dendrites of hippocampal pyramidal neurons}.
\bjtitle{Nature}
\bvolume{387}(\bissue{6636}),
\bfpage{869}--\blpage{875}
(\byear{1997})
\doiurl{10.1038/43119}
\end{barticle}
\endbibitem

\bibitem[\protect\citeauthoryear{Bi and Poo}{1998}]{Bi_Poo_1998}
\begin{botherref}
\oauthor{\bsnm{Bi}, \binits{G.-q.}},
\oauthor{\bsnm{Poo}, \binits{M.-m.}}:
Synaptic modifications in cultured hippocampal neurons: Dependence on spike timing, synaptic strength, and postsynaptic cell type.
The Journal of Neuroscience,
10464--10472
(1998)
\doiurl{10.1523/jneurosci.18-24-10464.1998}
\end{botherref}
\endbibitem

\bibitem[\protect\citeauthoryear{Rosenmund et~al.}{1995}]{Rosenmund_Feltz_Westbrook_1995}
\begin{barticle}
\bauthor{\bsnm{Rosenmund}, \binits{C.}},
\bauthor{\bsnm{Feltz}, \binits{A.}},
\bauthor{\bsnm{Westbrook}, \binits{G.L.}}:
\batitle{Calcium-dependent inactivation of synaptic nmda receptors in hippocampal neurons}.
\bjtitle{Journal of Neurophysiology}
\bvolume{73}(\bissue{1}),
\bfpage{427}--\blpage{430}
(\byear{1995})
\doiurl{10.1152/jn.1995.73.1.427}
\end{barticle}
\endbibitem

\bibitem[\protect\citeauthoryear{Zhigulin and Rabinovich}{2004}]{Zhigulin2004AnIR}
\begin{barticle}
\bauthor{\bsnm{Zhigulin}, \binits{V.P.}},
\bauthor{\bsnm{Rabinovich}, \binits{M.I.}}:
\batitle{An important role of spike timing dependent synaptic plasticity in the formation of synchronized neural ensembles}.
\bjtitle{Neurocomputing}
\bvolume{58-60},
\bfpage{373}--\blpage{378}
(\byear{2004})
\end{barticle}
\endbibitem

\bibitem[\protect\citeauthoryear{i~Petit and Murray}{2004}]{BofilliPetit2004SynchronyDA}
\begin{barticle}
\bauthor{\bsnm{Bofill-i-Petit}, \binits{A.}},
\bauthor{\bsnm{Murray}, \binits{A.F.}}:
\batitle{Synchrony detection and amplification by silicon neurons with stdp synapses}.
\bjtitle{IEEE Transactions on Neural Networks}
\bvolume{15},
\bfpage{1296}--\blpage{1304}
(\byear{2004})
\end{barticle}
\endbibitem

\bibitem[\protect\citeauthoryear{Toyoizumi et~al.}{2004}]{Toyoizumi2004SpiketimingDP}
\begin{bchapter}
\bauthor{\bsnm{Toyoizumi}, \binits{T.}},
\bauthor{\bsnm{Pfister}, \binits{J.-P.}},
\bauthor{\bsnm{Aihara}, \binits{K.}},
\bauthor{\bsnm{Gerstner}, \binits{W.}}:
\bctitle{Spike-timing dependent plasticity and mutual information maximization for a spiking neuron model}.
In: \bbtitle{Neural Information Processing Systems}
(\byear{2004})
\end{bchapter}
\endbibitem

\bibitem[\protect\citeauthoryear{Masquelier et~al.}{2008}]{Masquelier2008SpikeTD}
\begin{botherref}
\oauthor{\bsnm{Masquelier}, \binits{T.}},
\oauthor{\bsnm{Guyonneau}, \binits{R.}},
\oauthor{\bsnm{Thorpe}, \binits{S.J.}}:
Spike timing dependent plasticity finds the start of repeating patterns in continuous spike trains.
PLoS ONE
\textbf{3}
(2008)
\end{botherref}
\endbibitem

\bibitem[\protect\citeauthoryear{Masquelier et~al.}{2009}]{Masquelier2009CompetitiveSS}
\begin{barticle}
\bauthor{\bsnm{Masquelier}, \binits{T.}},
\bauthor{\bsnm{Guyonneau}, \binits{R.}},
\bauthor{\bsnm{Thorpe}, \binits{S.J.}}:
\batitle{Competitive stdp-based spike pattern learning}.
\bjtitle{Neural Computation}
\bvolume{21},
\bfpage{1259}--\blpage{1276}
(\byear{2009})
\end{barticle}
\endbibitem

\bibitem[\protect\citeauthoryear{Beyeler et~al.}{2013}]{Beyeler2013CategorizationAD}
\begin{barticle}
\bauthor{\bsnm{Beyeler}, \binits{M.}},
\bauthor{\bsnm{Dutt}, \binits{N.D.}},
\bauthor{\bsnm{Krichmar}, \binits{J.L.}}:
\batitle{Categorization and decision-making in a neurobiologically plausible spiking network using a stdp-like learning rule}.
\bjtitle{Neural networks : the official journal of the International Neural Network Society}
\bvolume{48},
\bfpage{109}--\blpage{24}
(\byear{2013})
\end{barticle}
\endbibitem

\bibitem[\protect\citeauthoryear{Masquelier and Thorpe}{2007}]{Masquelier2007UnsupervisedLO}
\begin{botherref}
\oauthor{\bsnm{Masquelier}, \binits{T.}},
\oauthor{\bsnm{Thorpe}, \binits{S.J.}}:
Unsupervised learning of visual features through spike timing dependent plasticity.
PLoS Computational Biology
\textbf{3}
(2007)
\end{botherref}
\endbibitem

\bibitem[\protect\citeauthoryear{Kheradpisheh et~al.}{2015}]{Kheradpisheh2015BioinspiredUL}
\begin{barticle}
\bauthor{\bsnm{Kheradpisheh}, \binits{S.R.}},
\bauthor{\bsnm{Ganjtabesh}, \binits{M.}},
\bauthor{\bsnm{Masquelier}, \binits{T.}}:
\batitle{Bio-inspired unsupervised learning of visual features leads to robust invariant object recognition}.
\bjtitle{Neurocomputing}
\bvolume{205},
\bfpage{382}--\blpage{392}
(\byear{2015})
\end{barticle}
\endbibitem

\bibitem[\protect\citeauthoryear{Diehl and Cook}{2015}]{Diehl2015UnsupervisedLO}
\begin{botherref}
\oauthor{\bsnm{Diehl}, \binits{P.U.}},
\oauthor{\bsnm{Cook}, \binits{M.}}:
Unsupervised learning of digit recognition using spike-timing-dependent plasticity.
Frontiers in Computational Neuroscience
\textbf{9}
(2015)
\end{botherref}
\endbibitem

\bibitem[\protect\citeauthoryear{Querlioz et~al.}{2013}]{Querlioz2013ImmunityTD}
\begin{barticle}
\bauthor{\bsnm{Querlioz}, \binits{D.}},
\bauthor{\bsnm{Bichler}, \binits{O.}},
\bauthor{\bsnm{Dollfus}, \binits{P.}},
\bauthor{\bsnm{Gamrat}, \binits{C.}}:
\batitle{Immunity to device variations in a spiking neural network with memristive nanodevices}.
\bjtitle{IEEE Transactions on Nanotechnology}
\bvolume{12},
\bfpage{288}--\blpage{295}
(\byear{2013})
\end{barticle}
\endbibitem

\bibitem[\protect\citeauthoryear{Kheradpisheh et~al.}{2016}]{Kheradpisheh2016STDPbasedSD}
\begin{barticle}
\bauthor{\bsnm{Kheradpisheh}, \binits{S.R.}},
\bauthor{\bsnm{Ganjtabesh}, \binits{M.}},
\bauthor{\bsnm{Thorpe}, \binits{S.J.}},
\bauthor{\bsnm{Masquelier}, \binits{T.}}:
\batitle{Stdp-based spiking deep neural networks for object recognition}.
\bjtitle{Neural networks : the official journal of the International Neural Network Society}
\bvolume{99},
\bfpage{56}--\blpage{67}
(\byear{2016})
\end{barticle}
\endbibitem

\bibitem[\protect\citeauthoryear{Ivanov and Michmizos}{2021}]{Ivanov2021IncreasingLS}
\begin{botherref}
\oauthor{\bsnm{Ivanov}, \binits{V.A.}},
\oauthor{\bsnm{Michmizos}, \binits{K.P.}}:
Increasing liquid state machine performance with edge-of-chaos dynamics organized by astrocyte-modulated plasticity.
ArXiv
\textbf{abs/2111.01760}
(2021)
\end{botherref}
\endbibitem

\bibitem[\protect\citeauthoryear{Tzounopoulos et~al.}{2007}]{Tzounopoulos_Rubio_Keen_Trussell_2007}
\begin{barticle}
\bauthor{\bsnm{Tzounopoulos}, \binits{T.}},
\bauthor{\bsnm{Rubio}, \binits{M.E.}},
\bauthor{\bsnm{Keen}, \binits{J.E.}},
\bauthor{\bsnm{Trussell}, \binits{L.O.}}:
\batitle{Coactivation of pre- and postsynaptic signaling mechanisms determines cell-specific spike-timing-dependent plasticity}.
\bjtitle{Neuron}
\bvolume{54}(\bissue{2}),
\bfpage{291}--\blpage{301}
(\byear{2007})
\doiurl{10.1016/j.neuron.2007.03.026}
\end{barticle}
\endbibitem

\bibitem[\protect\citeauthoryear{Softky and Koch}{1993}]{Softky_Koch_1993}
\begin{botherref}
\oauthor{\bsnm{Softky}, \binits{W.}},
\oauthor{\bsnm{Koch}, \binits{C.}}:
The highly irregular firing of cortical cells is inconsistent with temporal integration of random epsps.
The Journal of Neuroscience,
334--350
(1993)
\doiurl{10.1523/jneurosci.13-01-00334.1993}
\end{botherref}
\endbibitem

\bibitem[\protect\citeauthoryear{Rao and Sejnowski}{2001}]{Rao_Sejnowski_2001}
\begin{barticle}
\bauthor{\bsnm{Rao}, \binits{R.P.N.}},
\bauthor{\bsnm{Sejnowski}, \binits{T.J.}}:
\batitle{Spike-timing-dependent hebbian plasticity as temporal difference learning}.
\bjtitle{Neural Computation}
\bvolume{13}(\bissue{10}),
\bfpage{2221}--\blpage{2237}
(\byear{2001})
\doiurl{10.1162/089976601750541787}
\end{barticle}
\endbibitem

\bibitem[\protect\citeauthoryear{Bear and Singer}{1986}]{Bear_Singer_1986}
\begin{barticle}
\bauthor{\bsnm{Bear}, \binits{M.F.}},
\bauthor{\bsnm{Singer}, \binits{W.}}:
\batitle{Modulation of visual cortical plasticity by acetylcholine and noradrenaline}.
\bjtitle{Nature}
\bvolume{320}(\bissue{6058}),
\bfpage{172}--\blpage{176}
(\byear{1986})
\doiurl{10.1038/320172a0}
\end{barticle}
\endbibitem

\bibitem[\protect\citeauthoryear{Grossberg}{1987}]{Grossberg1987CompetitiveLF}
\begin{barticle}
\bauthor{\bsnm{Grossberg}, \binits{S.}}:
\batitle{Competitive learning: From interactive activation to adaptive resonance}.
\bjtitle{Cogn. Sci.}
\bvolume{11},
\bfpage{23}--\blpage{63}
(\byear{1987})
\end{barticle}
\endbibitem

\bibitem[\protect\citeauthoryear{N{\o}kland}{2016}]{Nkland2016DirectFA}
\begin{bchapter}
\bauthor{\bsnm{N{\o}kland}, \binits{A.}}:
\bctitle{Direct feedback alignment provides learning in deep neural networks}.
In: \bbtitle{Neural Information Processing Systems}
(\byear{2016})
\end{bchapter}
\endbibitem

\bibitem[\protect\citeauthoryear{Frenkel et~al.}{2021}]{Frenkel2021LearningWF}
\begin{botherref}
\oauthor{\bsnm{Frenkel}, \binits{C.}},
\oauthor{\bsnm{Lefebvre}, \binits{M.}},
\oauthor{\bsnm{Bol}, \binits{D.}}:
Learning without feedback: Fixed random learning signals allow for feedforward training of deep neural networks.
Frontiers in Neuroscience
\textbf{15}
(2021)
\end{botherref}
\endbibitem

\bibitem[\protect\citeauthoryear{Liao et~al.}{2015}]{Liao2015HowII}
\begin{botherref}
\oauthor{\bsnm{Liao}, \binits{Q.}},
\oauthor{\bsnm{Leibo}, \binits{J.Z.}},
\oauthor{\bsnm{Poggio}, \binits{T.A.}}:
How important is weight symmetry in backpropagation?
ArXiv
\textbf{abs/1510.05067}
(2015)
\end{botherref}
\endbibitem

\bibitem[\protect\citeauthoryear{Bartunov et~al.}{2018}]{Bartunov2018AssessingTS}
\begin{bchapter}
\bauthor{\bsnm{Bartunov}, \binits{S.}},
\bauthor{\bsnm{Santoro}, \binits{A.}},
\bauthor{\bsnm{Richards}, \binits{B.A.}},
\bauthor{\bsnm{Hinton}, \binits{G.E.}},
\bauthor{\bsnm{Lillicrap}, \binits{T.P.}}:
\bctitle{Assessing the scalability of biologically-motivated deep learning algorithms and architectures}.
In: \bbtitle{Neural Information Processing Systems}
(\byear{2018})
\end{bchapter}
\endbibitem

\bibitem[\protect\citeauthoryear{Grossberg}{1987}]{Grossberg_1987}
\begin{botherref}
\oauthor{\bsnm{Grossberg}, \binits{S.}}:
Competitive learning: From interactive activation to adaptive resonance.
Cognitive Science,
23--63
(1987)
\doiurl{10.1111/j.1551-6708.1987.tb00862.x}
\end{botherref}
\endbibitem

\bibitem[\protect\citeauthoryear{Lillicrap et~al.}{2020}]{Lillicrap_Santoro_Marris_Akerman_Hinton_2020}
\begin{botherref}
\oauthor{\bsnm{Lillicrap}, \binits{T.P.}},
\oauthor{\bsnm{Santoro}, \binits{A.}},
\oauthor{\bsnm{Marris}, \binits{L.}},
\oauthor{\bsnm{Akerman}, \binits{C.J.}},
\oauthor{\bsnm{Hinton}, \binits{G.}}:
Backpropagation and the brain.
Nature Reviews Neuroscience,
335--346
(2020)
\doiurl{10.1038/s41583-020-0277-3}
\end{botherref}
\endbibitem

\bibitem[\protect\citeauthoryear{Bengio}{2014}]{Bengio_2014}
\begin{botherref}
\oauthor{\bsnm{Bengio}, \binits{Y.}}:
How auto-encoders could provide credit assignment in deep networks via target propagation.
arXiv: Learning,arXiv: Learning
(2014)
\end{botherref}
\endbibitem

\bibitem[\protect\citeauthoryear{Lee et~al.}{2014}]{Lee_Zhang_Fischer_Bengio_2014}
\begin{botherref}
\oauthor{\bsnm{Lee}, \binits{D.-H.}},
\oauthor{\bsnm{Zhang}, \binits{S.}},
\oauthor{\bsnm{Fischer}, \binits{A.}},
\oauthor{\bsnm{Bengio}, \binits{Y.}}:
Difference target propagation.
ECML/PKDD
(2014)
\end{botherref}
\endbibitem

\bibitem[\protect\citeauthoryear{Bartunov et~al.}{2018}]{Bartunov_Santoro_Richards_Marris_Hinton_Lillicrap_2018}
\begin{botherref}
\oauthor{\bsnm{Bartunov}, \binits{S.}},
\oauthor{\bsnm{Santoro}, \binits{A.}},
\oauthor{\bsnm{Richards}, \binits{B.}},
\oauthor{\bsnm{Marris}, \binits{L.}},
\oauthor{\bsnm{Hinton}, \binits{G.}},
\oauthor{\bsnm{Lillicrap}, \binits{T.}}:
Assessing the scalability of biologically-motivated deep learning algorithms and architectures.
NIPS
(2018)
\end{botherref}
\endbibitem

\bibitem[\protect\citeauthoryear{Friston}{2008}]{friston2008hierarchical}
\begin{barticle}
\bauthor{\bsnm{Friston}, \binits{K.}}:
\batitle{Hierarchical models in the brain}.
\bjtitle{PLoS computational biology}
\bvolume{4}(\bissue{11}),
\bfpage{1000211}
(\byear{2008})
\end{barticle}
\endbibitem

\bibitem[\protect\citeauthoryear{Whittington and Bogacz}{2017}]{whittington2017approximation}
\begin{barticle}
\bauthor{\bsnm{Whittington}, \binits{J.C.}},
\bauthor{\bsnm{Bogacz}, \binits{R.}}:
\batitle{An approximation of the error backpropagation algorithm in a predictive coding network with local hebbian synaptic plasticity}.
\bjtitle{Neural computation}
\bvolume{29}(\bissue{5}),
\bfpage{1229}--\blpage{1262}
(\byear{2017})
\end{barticle}
\endbibitem

\bibitem[\protect\citeauthoryear{Millidge et~al.}{2022}]{millidge2022predictive}
\begin{barticle}
\bauthor{\bsnm{Millidge}, \binits{B.}},
\bauthor{\bsnm{Tschantz}, \binits{A.}},
\bauthor{\bsnm{Buckley}, \binits{C.L.}}:
\batitle{Predictive coding approximates backprop along arbitrary computation graphs}.
\bjtitle{Neural Computation}
\bvolume{34}(\bissue{6}),
\bfpage{1329}--\blpage{1368}
(\byear{2022})
\end{barticle}
\endbibitem

\bibitem[\protect\citeauthoryear{Buckley et~al.}{2017}]{buckley2017free}
\begin{barticle}
\bauthor{\bsnm{Buckley}, \binits{C.L.}},
\bauthor{\bsnm{Kim}, \binits{C.S.}},
\bauthor{\bsnm{McGregor}, \binits{S.}},
\bauthor{\bsnm{Seth}, \binits{A.K.}}:
\batitle{The free energy principle for action and perception: A mathematical review}.
\bjtitle{Journal of Mathematical Psychology}
\bvolume{81},
\bfpage{55}--\blpage{79}
(\byear{2017})
\end{barticle}
\endbibitem

\bibitem[\protect\citeauthoryear{Han et~al.}{2018}]{han2018deep}
\begin{botherref}
\oauthor{\bsnm{Han}, \binits{K.}},
\oauthor{\bsnm{Wen}, \binits{H.}},
\oauthor{\bsnm{Zhang}, \binits{Y.}},
\oauthor{\bsnm{Fu}, \binits{D.}},
\oauthor{\bsnm{Culurciello}, \binits{E.}},
\oauthor{\bsnm{Liu}, \binits{Z.}}:
Deep predictive coding network with local recurrent processing for object recognition.
Advances in neural information processing systems
\textbf{31}
(2018)
\end{botherref}
\endbibitem

\bibitem[\protect\citeauthoryear{Ororbia and Kifer}{2022}]{ororbia2022neural}
\begin{barticle}
\bauthor{\bsnm{Ororbia}, \binits{A.}},
\bauthor{\bsnm{Kifer}, \binits{D.}}:
\batitle{The neural coding framework for learning generative models}.
\bjtitle{Nature communications}
\bvolume{13}(\bissue{1}),
\bfpage{2064}
(\byear{2022})
\end{barticle}
\endbibitem

\bibitem[\protect\citeauthoryear{Baydin et~al.}{2022}]{baydin2022gradients}
\begin{botherref}
\oauthor{\bsnm{Baydin}, \binits{A.G.}},
\oauthor{\bsnm{Pearlmutter}, \binits{B.A.}},
\oauthor{\bsnm{Syme}, \binits{D.}},
\oauthor{\bsnm{Wood}, \binits{F.}},
\oauthor{\bsnm{Torr}, \binits{P.}}:
Gradients without backpropagation.
arXiv preprint arXiv:2202.08587
(2022)
\end{botherref}
\endbibitem

\bibitem[\protect\citeauthoryear{Silver et~al.}{2021}]{silver2021learning}
\begin{bchapter}
\bauthor{\bsnm{Silver}, \binits{D.}},
\bauthor{\bsnm{Goyal}, \binits{A.}},
\bauthor{\bsnm{Danihelka}, \binits{I.}},
\bauthor{\bsnm{Hessel}, \binits{M.}},
\bauthor{\bsnm{Hasselt}, \binits{H.}}:
\bctitle{Learning by directional gradient descent}.
In: \bbtitle{International Conference on Learning Representations}
(\byear{2021})
\end{bchapter}
\endbibitem

\bibitem[\protect\citeauthoryear{Wengert}{1964}]{wengert1964simple}
\begin{barticle}
\bauthor{\bsnm{Wengert}, \binits{R.E.}}:
\batitle{A simple automatic derivative evaluation program}.
\bjtitle{Communications of the ACM}
\bvolume{7}(\bissue{8}),
\bfpage{463}--\blpage{464}
(\byear{1964})
\end{barticle}
\endbibitem

\bibitem[\protect\citeauthoryear{Ren et~al.}{2022}]{ren2022scaling}
\begin{botherref}
\oauthor{\bsnm{Ren}, \binits{M.}},
\oauthor{\bsnm{Kornblith}, \binits{S.}},
\oauthor{\bsnm{Liao}, \binits{R.}},
\oauthor{\bsnm{Hinton}, \binits{G.}}:
Scaling forward gradient with local losses.
arXiv preprint arXiv:2210.03310
(2022)
\end{botherref}
\endbibitem

\bibitem[\protect\citeauthoryear{Bengio et~al.}{2006}]{BengioLPL06}
\begin{bchapter}
\bauthor{\bsnm{Bengio}, \binits{Y.}},
\bauthor{\bsnm{Lamblin}, \binits{P.}},
\bauthor{\bsnm{Popovici}, \binits{D.}},
\bauthor{\bsnm{Larochelle}, \binits{H.}}:
\bctitle{Greedy layer-wise training of deep networks}.
In: \beditor{\bsnm{Sch{\"{o}}lkopf}, \binits{B.}},
\beditor{\bsnm{Platt}, \binits{J.C.}},
\beditor{\bsnm{Hofmann}, \binits{T.}} (eds.)
\bbtitle{Advances in Neural Information Processing Systems 19, Proceedings of the Twentieth Annual Conference on Neural Information Processing Systems, Vancouver, British Columbia, Canada, December 4-7, 2006},
pp. \bfpage{153}--\blpage{160}.
\bpublisher{{MIT} Press}, \blocation{???}
(\byear{2006})
\end{bchapter}
\endbibitem

\bibitem[\protect\citeauthoryear{Scellier and Bengio}{2019}]{scellier2019equivalence}
\begin{barticle}
\bauthor{\bsnm{Scellier}, \binits{B.}},
\bauthor{\bsnm{Bengio}, \binits{Y.}}:
\batitle{Equivalence of equilibrium propagation and recurrent backpropagation}.
\bjtitle{Neural computation}
\bvolume{31}(\bissue{2}),
\bfpage{312}--\blpage{329}
(\byear{2019})
\end{barticle}
\endbibitem

\bibitem[\protect\citeauthoryear{Laborieux et~al.}{2021}]{laborieux2021scaling}
\begin{barticle}
\bauthor{\bsnm{Laborieux}, \binits{A.}},
\bauthor{\bsnm{Ernoult}, \binits{M.}},
\bauthor{\bsnm{Scellier}, \binits{B.}},
\bauthor{\bsnm{Bengio}, \binits{Y.}},
\bauthor{\bsnm{Grollier}, \binits{J.}},
\bauthor{\bsnm{Querlioz}, \binits{D.}}:
\batitle{Scaling equilibrium propagation to deep convnets by drastically reducing its gradient estimator bias}.
\bjtitle{Frontiers in neuroscience}
\bvolume{15},
\bfpage{633674}
(\byear{2021})
\end{barticle}
\endbibitem

\bibitem[\protect\citeauthoryear{O’Connor et~al.}{2019}]{o2019training}
\begin{bchapter}
\bauthor{\bsnm{O’Connor}, \binits{P.}},
\bauthor{\bsnm{Gavves}, \binits{E.}},
\bauthor{\bsnm{Welling}, \binits{M.}}:
\bctitle{Training a spiking neural network with equilibrium propagation}.
In: \bbtitle{The 22nd International Conference on Artificial Intelligence and Statistics},
pp. \bfpage{1516}--\blpage{1523}
(\byear{2019}).
\bcomment{PMLR}
\end{bchapter}
\endbibitem

\bibitem[\protect\citeauthoryear{Fang et~al.}{2023}]{fang2023spikingjelly}
\begin{barticle}
\bauthor{\bsnm{Fang}, \binits{W.}},
\bauthor{\bsnm{Chen}, \binits{Y.}},
\bauthor{\bsnm{Ding}, \binits{J.}},
\bauthor{\bsnm{Yu}, \binits{Z.}},
\bauthor{\bsnm{Masquelier}, \binits{T.}},
\bauthor{\bsnm{Chen}, \binits{D.}},
\bauthor{\bsnm{Huang}, \binits{L.}},
\bauthor{\bsnm{Zhou}, \binits{H.}},
\bauthor{\bsnm{Li}, \binits{G.}},
\bauthor{\bsnm{Tian}, \binits{Y.}}:
\batitle{Spikingjelly: An open-source machine learning infrastructure platform for spike-based intelligence}.
\bjtitle{Science Advances}
\bvolume{9}(\bissue{40}),
\bfpage{1480}
(\byear{2023})
\end{barticle}
\endbibitem

\bibitem[\protect\citeauthoryear{Maass}{1997}]{Maas1997NetworksOS}
\begin{barticle}
\bauthor{\bsnm{Maass}, \binits{W.}}:
\batitle{Networks of spiking neurons: the third generation of neural network models}.
\bjtitle{Neural Networks}
\bvolume{14},
\bfpage{1659}--\blpage{1671}
(\byear{1997})
\end{barticle}
\endbibitem

\bibitem[\protect\citeauthoryear{Akopyan et~al.}{2015}]{akopyan2015truenorth}
\begin{barticle}
\bauthor{\bsnm{Akopyan}, \binits{F.}},
\bauthor{\bsnm{Sawada}, \binits{J.}},
\bauthor{\bsnm{Cassidy}, \binits{A.}},
\bauthor{\bsnm{Alvarez-Icaza}, \binits{R.}},
\bauthor{\bsnm{Arthur}, \binits{J.}},
\bauthor{\bsnm{Merolla}, \binits{P.}},
\bauthor{\bsnm{Imam}, \binits{N.}},
\bauthor{\bsnm{Nakamura}, \binits{Y.}},
\bauthor{\bsnm{Datta}, \binits{P.}},
\bauthor{\bsnm{Nam}, \binits{G.-J.}}, \betal:
\batitle{Truenorth: Design and tool flow of a 65 mw 1 million neuron programmable neurosynaptic chip}.
\bjtitle{IEEE transactions on computer-aided design of integrated circuits and systems}
\bvolume{34}(\bissue{10}),
\bfpage{1537}--\blpage{1557}
(\byear{2015})
\end{barticle}
\endbibitem

\bibitem[\protect\citeauthoryear{Davies et~al.}{2018}]{davies2018loihi}
\begin{barticle}
\bauthor{\bsnm{Davies}, \binits{M.}},
\bauthor{\bsnm{Srinivasa}, \binits{N.}},
\bauthor{\bsnm{Lin}, \binits{T.-H.}},
\bauthor{\bsnm{Chinya}, \binits{G.}},
\bauthor{\bsnm{Cao}, \binits{Y.}},
\bauthor{\bsnm{Choday}, \binits{S.H.}},
\bauthor{\bsnm{Dimou}, \binits{G.}},
\bauthor{\bsnm{Joshi}, \binits{P.}},
\bauthor{\bsnm{Imam}, \binits{N.}},
\bauthor{\bsnm{Jain}, \binits{S.}}, \betal:
\batitle{Loihi: A neuromorphic manycore processor with on-chip learning}.
\bjtitle{Ieee Micro}
\bvolume{38}(\bissue{1}),
\bfpage{82}--\blpage{99}
(\byear{2018})
\end{barticle}
\endbibitem

\end{thebibliography}






\end{document}